\documentclass[lettersize,journal]{IEEEtran}

\usepackage{amsmath}
\usepackage{booktabs}

\usepackage{algpseudocode} 
\usepackage{algorithm}
\algrenewcommand\algorithmicrequire{\textbf{Input:}}
\algrenewcommand\algorithmicensure{\textbf{Output:}}

\usepackage{amsmath,amsfonts}

\usepackage[font=tiny]{subcaption}

\usepackage{array}
\usepackage[caption=false,font=footnotesize,labelfont=sf,textfont=sf]{subfig}
\usepackage{textcomp}
\usepackage{stfloats}
\usepackage{url}
\usepackage{verbatim}
\usepackage{graphicx}
\usepackage{cite}
\usepackage{cuted}

\usepackage{multirow}
\usepackage{makecell}
\usepackage{multicol}
\usepackage{bbm}
\usepackage{xcolor}
\usepackage{xspace}
\usepackage{float}
\usepackage{color}
\usepackage{colortbl}

\usepackage{amsmath,bm}

\usepackage{times}
\usepackage{epsfig}

\usepackage{mathrsfs}
\usepackage[switch]{lineno}

\usepackage{caption}
\definecolor{demphcolor}{RGB}{144,144,144}
\newcommand{\demph}[1]{\textcolor{demphcolor}{#1}}

\newcommand{\tablestyle}[2]{\setlength{\tabcolsep}{#1}\renewcommand{\arraystretch}{#2}\centering\footnotesize}
\usepackage{makecell}
\usepackage[export]{adjustbox}
\usepackage[normalem]{ulem}
\usepackage{textpos}  %
\usepackage{color}
\newcommand{\dt}[1]{\fontsize{5pt}{0.1em}\selectfont (#1)}

\newlength\savewidth\newcommand\shline{\noalign{\global\savewidth\arrayrulewidth
  \global\arrayrulewidth 1pt}\hline\noalign{\global\arrayrulewidth\savewidth}}

\makeatletter\renewcommand\paragraph{\@startsection{paragraph}{4}{\z@}
  {.5em \@plus1ex \@minus.2ex}{-.5em}{\normalfont\normalsize\bfseries}}\makeatother

\newcolumntype{I}{!{\vrule width 3pt}}
\newlength\savedwidth

\def\eg{\emph{e.g., }}

\def\ie{\emph{i.e., }}

\newcommand{\cisdq}{\textsc{\textbf{C}i\textbf{SDQ}}\xspace}

\definecolor{green}{rgb}{1,0,0}

\newcommand{\hy}{\hat{\bm{y}}}

\renewcommand{\Sigma}{\mathfrak{S}}

\newcommand{\lmatch}[1]{{\mathcal{L}}_{\rm match}(#1)}

\def\eqref#1{equation~\ref{#1}}

\def\1{\bm{1}}

\def\vc{{\bm{c}}}

\def\vm{{\bm{m}}}

\def\vq{{\bm{q}}}

\def\vx{{\bm{x}}}

\DeclareMathAlphabet{\mathsfit}{\encodingdefault}{\sfdefault}{m}{sl}
\SetMathAlphabet{\mathsfit}{bold}{\encodingdefault}{\sfdefault}{bx}{n}

\DeclareMathOperator*{\argmin}{arg\,min}

\begin{document}


\title{Continual Learning for Image Segmentation with Dynamic Query}

\author{Weijia Wu$^*$, Yuzhong Zhao$^*$,
Zhuang Li,  Lianlei Shan, 
Hong Zhou$^\dagger$, Mike Zheng Shou
\thanks{W. Wu and H. Zhou are with the Zhejiang University, Hangzhou, 310058, China. (e-mail: weijiawu@zju.edu.cn; zhouh@mail.bme.zju.edu.cn).}
\thanks{Y. Zhao and L. Shan are with the School of Computer Science and Technology, University of Chinese Academy of Sciences, Beijing, 101408, China.}
\thanks{Z. Li is with the Kuaishou Technology, Beijing, China.}
\thanks{M. Shou is with the National University of Singapore, Singapore.}
\thanks{$*$Equal contribution.}
\thanks{$\dagger$Corresponding author (Z. Hong).}
\thanks{Copyright © 2023 IEEE. Personal use of this material is permitted. However, permission to use this material for any other purposes must be obtained from the IEEE by sending an email to pubs-permissions\@ieee.org.}
}


\markboth{IEEE Transactions on Circuits and Systems for Video Technology}%
{Continual Learning for Image Segmentation with Dynamic Query}

\maketitle

\begin{abstract}
Image segmentation based on continual learning 
exhibits a critical drop of performance, mainly due to catastrophic forgetting and background shift, 
as 
they are required to incorporate new classes continually.
In this paper, we propose a simple, yet 
effective 
\textbf{C}ontinual \textbf{I}mage \textbf{S}egmentation method with incremental \textbf{D}ynamic \textbf{Q}uery~(\cisdq), which 
decouples the representation learning of both old and new knowledge with lightweight query embedding.
\cisdq mainly includes three contributions: %
1) We define \textit{dynamic queries} with adaptive background class to 
exploit 
past knowledge and learn future classes naturally. 
2) \cisdq  proposes a class/instance-aware Query Guided Knowledge Distillation strategy to overcome catastrophic forgetting 
by 
capturing the inter-class diversity and intra-class identity.
3) 
Apart from 
semantic segmentation, 
\cisdq 
introduce the continual learning for \textit{instance segmentation} in which instance-wise labeling and supervision are considered.
Extensive experiments on three datasets for two tasks~(\ie{} continual semantic and instance segmentation are conducted to demonstrate that \cisdq achieves the state-of-the-art performance, specifically, obtaining 4.4\% and 2.9\% mIoU improvements for the ADE 100-10 (6 steps) setting and ADE 100-5 (11 steps) setting.
%
\end{abstract}

\begin{IEEEkeywords}
Image Segmentation, Continual Learning, Dynamic Query, Dense matching strategy, Transformer.
\end{IEEEkeywords}

\section{Introduction}

\IEEEPARstart{I}{mage} segmentation, including semantic segmentation and instance segmentation, 
is a 
fundamental 
task in 
computer vision.
In recent years, data-driven segmentation networks~\cite{(deeplabv3)chen2017rethinking,cheng2022masked} have made extraordinary progress with fully-supervised learning of fixed data, where all classes are fixed and known beforehand and learned at once. 
However, in a real-world system, it is preferable that one model can dynamically update its knowledge and extend to segment new classes without retraining from scratch.
To reach the setup, previous works~\cite{(PLOP)douillard2020plop,zhang2022representation,cha2021ssul} propose to introduce the class incremental learning for semantic segmentation, named continual semantic segmentation~(CSS).
Different from these works that focus on semantic segmentation, in this paper, we try to establish a generic continual image segmentation task, which includes semantic segmentation and instance segmentation.

Continual learning-based image segmentation (or continual image segmentation, CIS) is a task that requires one model incrementally learn and segment newly arriving class objects while not catastrophically forgetting the past learned classes (old classes). 
Similar to continual semantic segmentation, continual image segmentation also faces two main challenges.
The first one is the \textbf{catastrophic forgetting}~\cite{robins1995catastrophic,french1999catastrophic}, where the model quickly fits the new data distribution and loses the old discriminative representation.
During continual learning, the new training data and class annotation is only used, and the old classes are not available usually be treated as background.
The network usually tends to catastrophically and abruptly forget previously learned knowledge~(old classes) when learning new information~(new classes). 
Existing continual semantic segmentation methods~\cite{(PLOP)douillard2020plop,cha2021ssul} typically adopt inefficient and explicit multiple strategies to tackle the problem.
For instance, PLOP~\cite{(PLOP)douillard2020plop} naively adopts a spatial distillation loss without fine-grained relationship learning of intra-class, inter-class, each pixel.
SSUL~\cite{cha2021ssul} adopts three strategies, \ie{} pseudo-labeling, exemplar memory, and model freezing to cover the challenge, but pseudo-labeling usually shows extremely instability and exemplar memory requires the extra cost memory of the network.

\begin{figure}
\centering
  \includegraphics[width=1.0\linewidth]{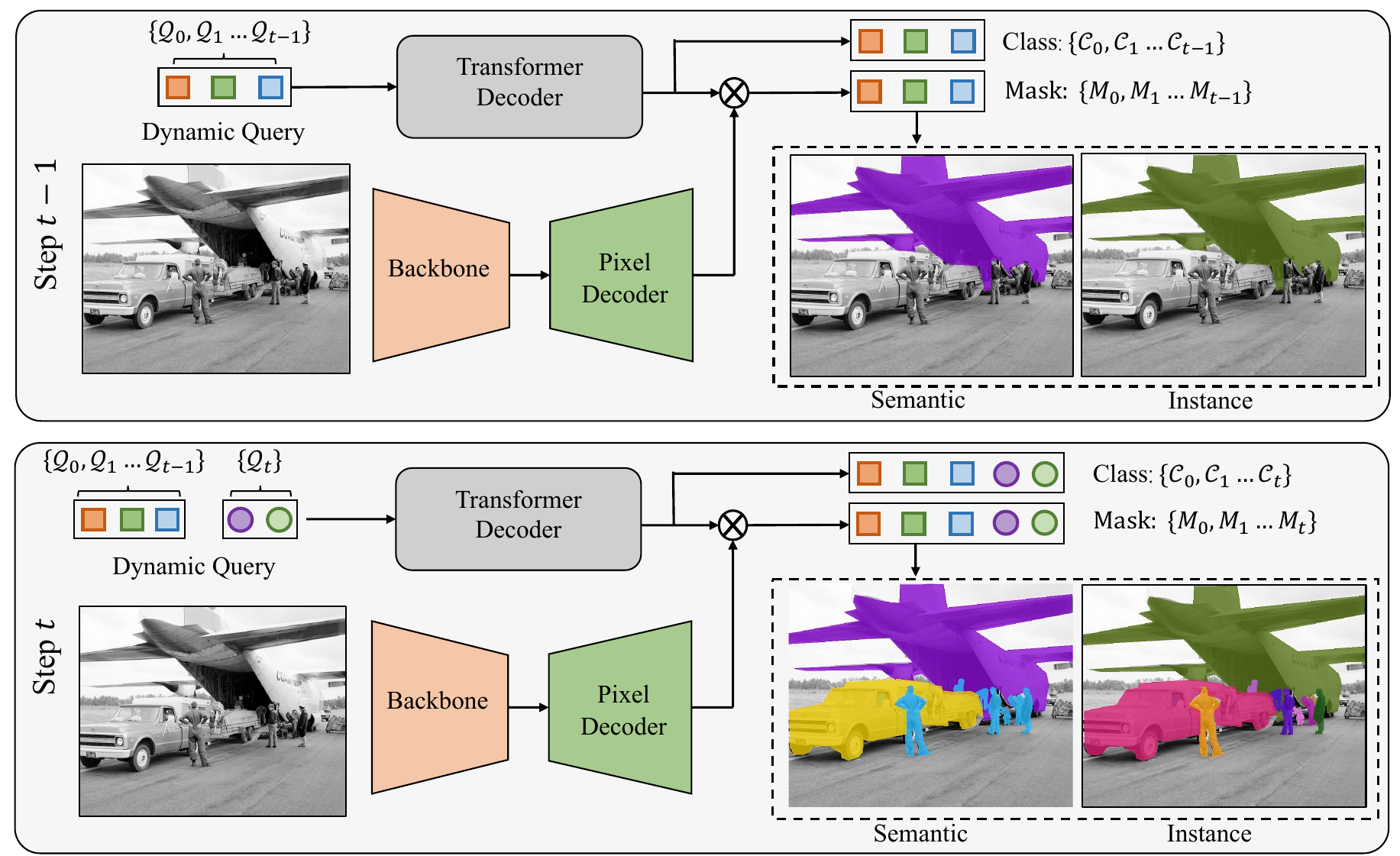}
  \caption{ \textbf{Illustration of \cisdq}. Dynamic query decouples the representation learning of old and new knowledge with dynamically increasing lightweight queries. $Q_{t}$ , $C_{t}$ and $M_{t}$ refers to the Query group, Classes, and Masks in $t$-th step, respectively.}
\label{fig:comparison}
\end{figure}

The second challenge, inherited from continual semantic segmentation, is the \textbf{background shift}~\cite{(PLOP)douillard2020plop,cha2021ssul}.
Different from the fixed and certain background pixels of traditional image segmentation, the background pixel for continual semantic segmentation belongs to \textit{three categories}: potential future object classes, past object classes, and the true background.
For instance, if an image contains three object classes, \ie{} \texttt{sofa}, \texttt{person}, \texttt{dog}, where only class \texttt{sofa} mask annotation is
available and class \texttt{person} and \texttt{dog} belong to the past classes without annotation.
If the model naively treats all unknown background pixels to the true background pixel class, the old and future knowledge will be further damaged and forgotten.
Some works~\cite{(PLOP)douillard2020plop,cha2021ssul} try to solve background shift with pseudo-labeling from the old model or other off-the-shelf saliency-map detectors. 
However, the predicted pseudo-labeling from the model usually present instability and worse performance, and it is impossible to predict all potential objects for future classes.

In this paper, we propose a simple, yet efficient \textbf{C}ontinual \textbf{I}mage \textbf{S}egmentation method with incremental \textbf{D}ynamic \textbf{Q}uery~(\cisdq), which 
decouples the representation learning of both old and new knowledge with class/instance-aware lightweight query embedding.
Inspired by query-based modeling for vision tasks~\cite{cheng2022masked,wu2022end,wu2021bilingual,zhao2023explore,gong2022curiosity}, we propose a new concept \textit{dynamic query}, enabling the model dynamically update its knowledge and extend to segment new classes via dynamically increasing query embedding of new classes or instances.
As shown in Fig.~\ref{fig:comparison}, we adopt one query embedding to represent one class or instance and dynamically increase the number of query embeddings to extend new classes without retraining from scratch.  
\cisdq main includes two advantages: 1) Different queries are used to represent different classes at each step to handle \textit{catastrophic forgetting} problem.
At step $t$, queries $\{Q_0,Q_1...Q_{t-1}\}$ will be used to retain the past knowledge of old classes~(\ie{} $\{\mathcal{C}_0,\mathcal{C}_1...\mathcal{C}_{t-1}\}$), newly added queries $\{Q_{t}\}$ are used to learn new appeared knowledge~(classes) $\{\mathcal{C}_{t}\}$.
2) \textit{Adaptive} background concept for each query of each step is proposed to solve \textit{background shift} problem. 
Different background definitions $\{\mathcal{B}_0,\mathcal{B}_1...\mathcal{B}_{t-1}\}$ for different queries $\{Q_0,Q_1...Q_{t-1}\}$ are used to cover different background representation at each step.
CNN-based multi-classification per-pixel in one feature map~\cite{(deeplabv3)chen2017rethinking} is transformed to a set of binary-classification in multi-maps~\cite{cheng2022masked}, thus there are $t$ different definitions of background for $t$ predicted masks each associated with a single category.
For instance, at step $t-1$, if only class \texttt{sofa} mask annotation is available,  \texttt{person}, \texttt{dog} pixel region as the \textit{unknown} region is defined as the background $\mathcal{B}_{t-1}$.
But at step $t$, \texttt{person} as newly added class is defined as the positive sample, but \texttt{sofa} mask is not available, belonging to the background class $\mathcal{B}_{t}$.
Thus the adaptive background does not give confusing supervision, while there is no background shift. 

Besides, we propose a query-guided knowledge distillation~(Query G-KD) to further alleviate catastrophic forgetting.
Different from previous knowledge distillation of semantic segmentation, Query G-KD can achieve more precise class/instance-aware knowledge distillation via capturing the inter-class diversity and intra-class identity, while each query is responsible for one class.
Similarly, the knowledge of inter-instance diversity and intra-instance identity can be retained with Query G-KD for continual instance segmentation tasks.
To 
summarize, our contributions are four-folds:
\begin{itemize}
    \item We propose a simple, yet efficient unified \textbf{C}ontinual \textbf{I}mage \textbf{S}egmentation method with incremental \textbf{D}ynamic \textbf{Q}uery, namely \cisdq, including a \textit{adaptive} background concept for each class/instance, which decouples the representation learning of old and new knowledge with dynamically increasing lightweight query embedding.
    
    \item \cisdq  introduces a Query Guided Knowledge Distillation~(Query G-KD) strategy to overcome catastrophic forgetting via capturing the inter-class diversity and intra-class identity.
    
    \item \cisdq introduces continual learning for \textit{instance segmentation} in which instance-wise labeling and supervision are considered. We also solve semantic and instance segmentation in a unified framework.
    
    \item Experiments are conducted on \textit{continual semantic segmentation} and \textit{continual instance segmentation} on three datasets, respectively. And \cisdq achieves state-of-the-art performance with up to
    \textbf{$10\%$} improvements than previous works. 
    
\end{itemize}

\section{Related Work}

\subsection{Image Segmentation}
Image Segmentation mainly includes two tasks: semantic and instance segmentation tasks.
For semantic segmentation, Some early works attempted~\cite{allili2010image,salgado2000efficient,ida1995image,lu2007binary,sun2003semiautomatic} to explore the use of certain image priors, such as Gaussian Mixture Modeling, to segment images or videos.
Fully Convolutional Networks~(FCN)~\cite{long2015fully} is the first deep learning based work to perform pixel-to-pixel semantic classification in an end-to-end manner.
After that, researchers focused on different aspects for improving semantic segmentation, \eg{} contextual relationships~\cite{arnab2016higher,zheng2015conditional,zhang2020causal,zhang2021self}, Spatial pyramid pooling~\cite{chen2014semantic,zhao2017pyramid,liu2015parsenet}, and image pyramid~\cite{eigen2015predicting,farabet2012learning}.
In recent years, several works~\cite{xie2021segformer,cheng2021per} try to adopt transformer architectures for semantic segmentation.
Segformer~\cite{xie2021segformer} design a novel hierarchically structured Transformer encoder to output multi-scale features.
For instance segmentation~\cite{hariharan2014simultaneous}, Mask R-CNN~\cite{he2017mask}, as the representation, extend Faster R-CNN~\cite{ren2015faster} by adding a branch for predicting an object mask in parallel with the existing branch for bounding box recognition. 
DETR~\cite{carion2020end} proposed to segment instances with mask attention, which is more natural.
Maskformer~\cite{cheng2021per} proposed mask classification modeling, which solves semantic- and instance-level segmentation tasks in a unified manner.
GenPromp~\cite{Zhao_2023_ICCV}, DiffuMask~\cite{wu2023diffumask} and DatasetDM~\cite{wu2023datasetdm} proposed to use diffusion model for enhancing semantic segmentation tasks, supporting open-set segmentation.
In this paper, we try to utilize the advantage of a transformer for continual image segmentation.

\subsection{Class Continual Learning}
Class continual learning method~\cite{yu2023contrastive,fu2023continual,le2022uifgan,zhou2021image,xu2020u2fusion,yan2020interactive,yan2020social} mainly focuses on the classification task, and alleviates catastrophic forgetting~\cite{mccloskey1989catastrophic}, which is caused by the domain distribution change from the training data.
To solve this problem, most works~\cite{bang2021rainbow,belouadah2019il2m,chaudhry2021using} try to maintain the performance of old classes with knowledge distillation~\cite{chaudhry2018riemannian,cheraghian2021semantic,douillard2020podnet}, and adversarial training~\cite{xiang2019incremental,ebrahimi2020adversarial}.
In recent years, inspired by the success of transformer architecture~\cite{dosovitskiy2020image,carion2020end} in computer vision, some works try to solve class incremental learning with transformer~\cite{xue2022meta,ashok2022class,douillard2022dytox,boschini2022transfer}.
MEta-ATtention~\cite{xue2022meta} proposed to use a pre-trained ViT to new tasks without sacrificing performance on already learned tasks. 
DyTox~\cite{douillard2022dytox} designs a shared transformer encoder and decoder to cover incremental learning with a dynamic expansion of special tokens.
TwF~\cite{boschini2022transfer} proposed a hybrid
approach building upon a fixed pre-trained sibling network, which continuously propagates the knowledge inherent from the previous task through a
layer-wise loss term. 
However, no one tries to adopt transformer to solve continual semantic and instance segmentation.
%

\subsection{Continual Image Segmentation}

\textbf{Semantic Segmentation.}
Except for catastrophic forgetting, continual semantic segmentation also faces another challenge, \ie{} background shift~\cite{(PLOP)douillard2020plop,cha2021ssul}.
Most existing methods solve the two challenges with rehearsal-based~\cite{huang2021half,yan2021framework}, pesudo label~\cite{cermelli2020modeling,(PLOP)douillard2020plop,zhao2022rbc,zheng2021continual}, and knowledge distillation~\cite{(PLOP)douillard2020plop,(MiB)cermelli2020modeling}.
MiB~\cite{(MiB)cermelli2020modeling} propose a new objective function and introduce a specific classifier initialization strategy to solve the background shift.
PLOP~\cite{(PLOP)douillard2020plop} proposed a multi-scale pooling distillation and entropy-based pseudo-labelling of the background to deal with the two problems.
RCIL~\cite{zhang2022representation} proposed a structural re-parameterization to decouple the representation learning of both old and new knowledge for solving catastrophic forgetting.
SSUL~\cite{cha2021ssul} solves background shift and catastrophic forgetting with three strategies, \ie{} unknown classes in background class, freeze backbone, and exemplar memory.
To avoid forgetting old
knowledge, MicroSeg~\cite{zhang2022mining} first splits the given image
into hundreds of segment proposals with a proposal generator. Those segment proposals with strong objectness from the background are then clustered and assigned
newly-defined labels during the optimization.
Based on Segformer~\cite{xie2021segformer}, SATS~\cite{qiu2022sats} design a knowledge distillation for transformer-based semantic segmentation.
However, the above method still can not achieve preferable performance, while they can not solve the two challenges entirely.
Different from previous methods, We propose an incremental dynamic query and query-guided knowledge distillation to decouple the representation learning of old and new knowledge.

\textbf{Instance Segmentation.} 
Despite enormous progress in the continual learning and instance segmentation tasks, almost no work try to solve continual problem in instance segmentation, while continual instance segmentation is also important 
%
%
And there exists a related task, \ie{} incremental few-shot instance segmentation. 
iMTFA~\cite{ganea2021incremental} design the first approach for incremental few-shot instance segmentation, which match these class embeddings at the RoI-level using cosine similarity.
iFS-RCNN~\cite{nguyen2022ifs} leverages Bayesian learning to address a paucity of training examples of new classes.
But the setting of incremental few-shot instance segmentation is different with continual instance segmentation in two parts: 1). Continual instance segmentation requires model to learn feature of new class during \textit{multi-steps}, while incremental few-shot instance segmentation only includes two-steps.
2). Continual instance segmentation provides abundant data of new classes, while incremental few-shot instance segmentation only training on a few data.
Besides, different from the above two methods, the proposed dynamic query and query-guided knowledge distillation can decouple the representation learning of old and new knowledge, which can deal with catastrophic forgetting and background shift challenges effectively. 

\begin{figure*}[t]
	\begin{minipage}{0.48\linewidth}
		\includegraphics[width=0.99\linewidth]{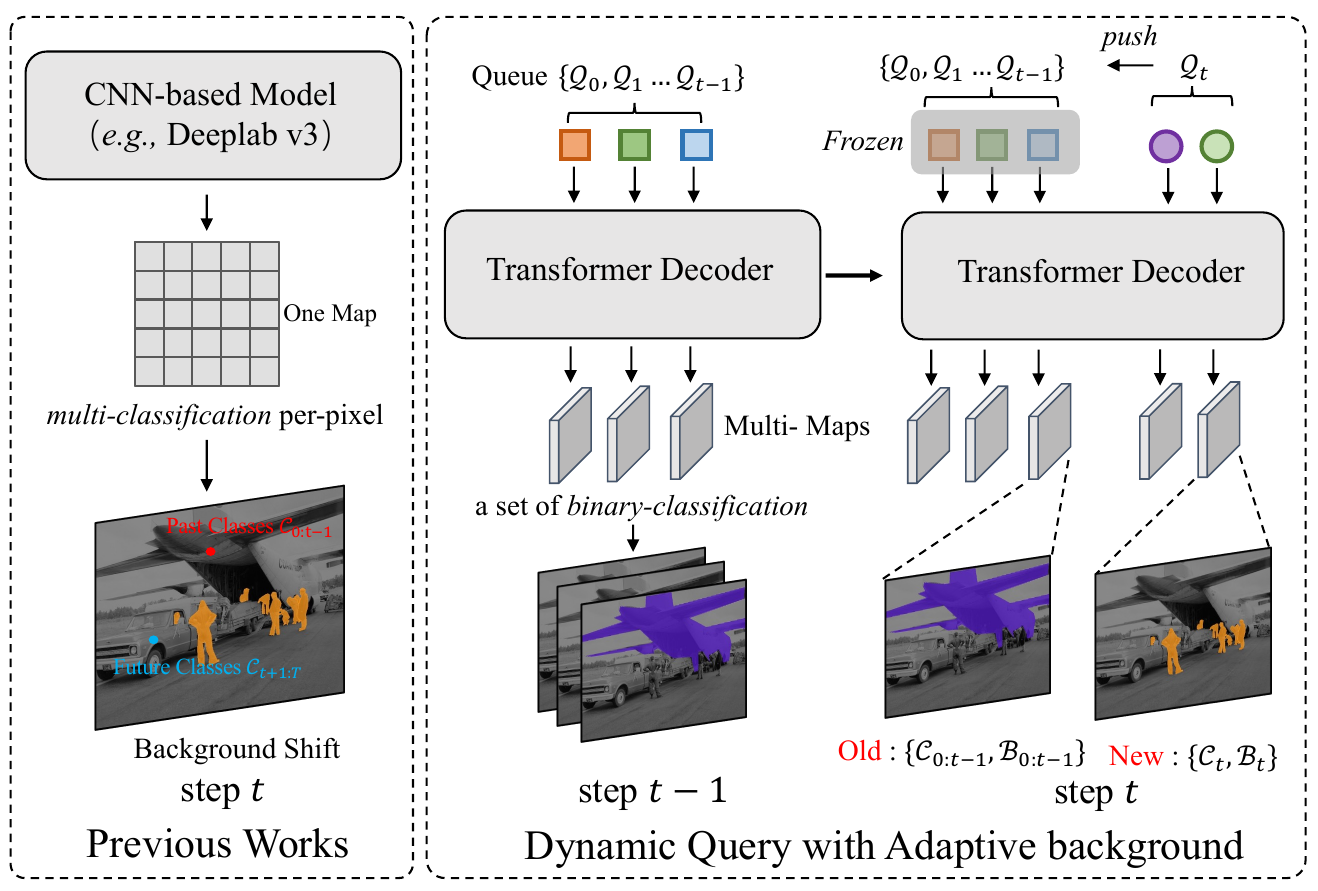}
		\subcaption{Dynamic Query for Continual Image Segmentation.}
		\label{fig:3a}	
	\end{minipage}
    \quad
	\begin{minipage}{0.49\linewidth}
		\centering
		\includegraphics[width=0.99\linewidth]{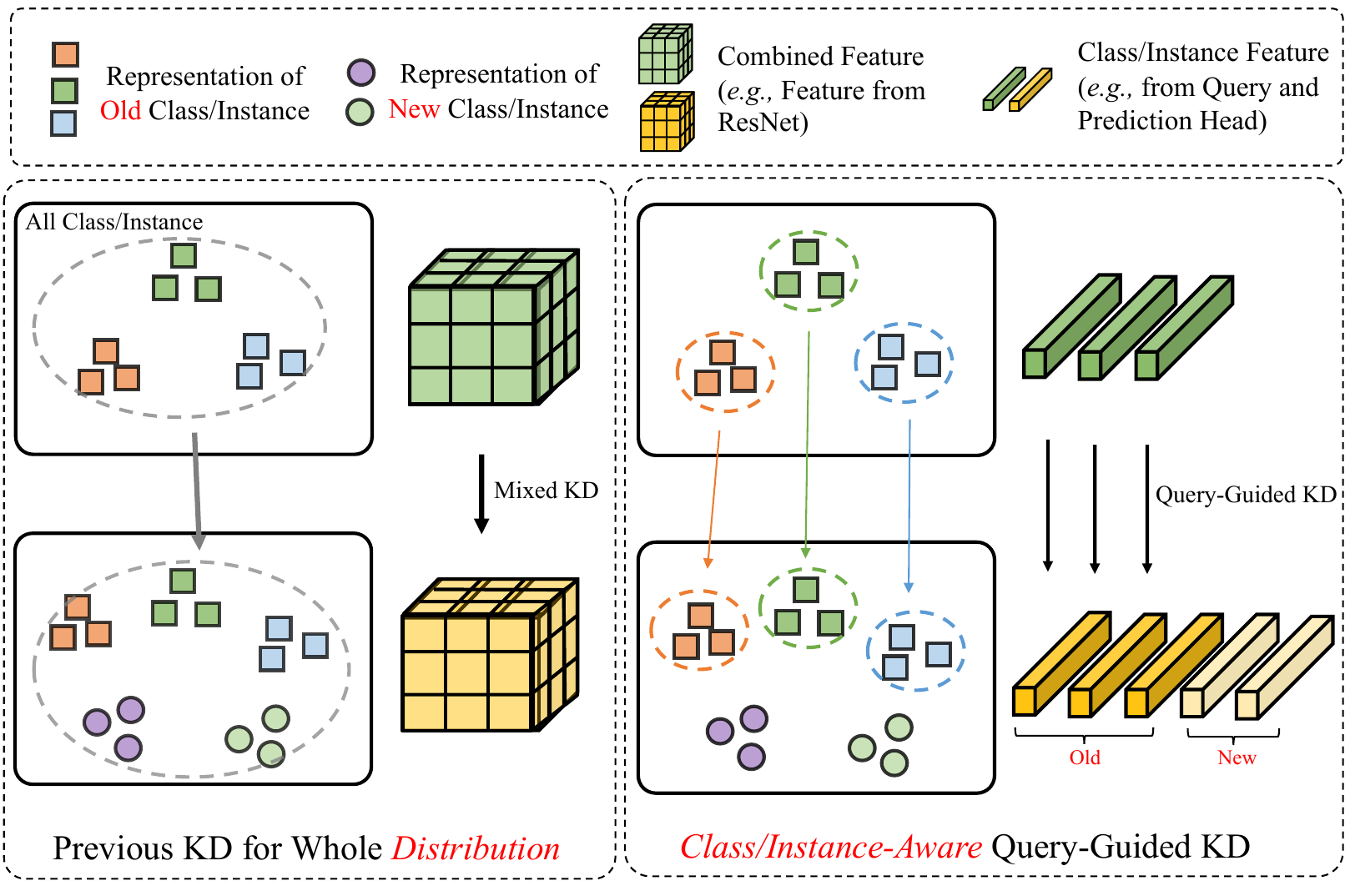}
		\subcaption{Class/Instance-Aware Query-Guided KD.}
		 \label{fig:3b}	
	\end{minipage}
	\caption{\textbf{Illustration for Dynamic Query and Query-Guided Knowledge Distillation.} (a) Incremental Dynamic Query decouples the representation learning of old and new knowledge with dynamically increasing lightlweight query embedding; (b) Compared with previous knowledge distillation~\cite{(PLOP)douillard2020plop,cha2021ssul,zhang2022representation} over the whole feature distribution, the proposed Query-Guided KD develop class/instance level distillation to overcome catastrophic forgetting challenge.}
\label{ablation_semantic_visual}
\end{figure*}

\section{Notations and Problem Setting}

Similar to previous works~\cite{(PLOP)douillard2020plop,cha2021ssul},  for the continual learning scenario, each training on the newly added classes and dataset as a \emph{step}, where existing $T$ steps.
For $t$-th step, given a base model $f^{t-1}_{\bm\theta}$ with parameter $\theta_{t-1}$ trained on $\{\mathcal{D}_0,\mathcal{D}_1...\mathcal{D}_{t-1} \}$ with $\{\mathcal{C}_0, \mathcal{C}_1...\mathcal{C}_{t-1}\}$ classes, the model is expected to segment $\sum_{i=0}^t \mathcal{C}_i$ classes after training on the newly added dataset $\mathcal{D}_t$ with extra $\mathcal{C}_t$ new classes, where the training data of old classes are not accessible. 
The network usually quickly fits the new data distribution with extra $\mathcal{C}_t$ new classes, and cause serious performance drop for old classes. 
The challenge in continual image segmentation is named \textit{catastrophic forgetting}.
In this paper, existing $T$ different query groups $\{Q_0,Q_1...Q_{T}\}$ are responsible for $\{\mathcal{C}_0, \mathcal{C}_1...\mathcal{C}_{T}\}$ classes at each step. 
Therefore, we can decouple the representation learning of old and new knowledge for step $t$ from two aspects: 1) Query groups $\{Q_0,Q_1...Q_{t-1}\}$ will be frozen to retain the past knowledge of old classes~(\ie{} $\sum_{i=0}^{t-1} \mathcal{C}_i$), newly added queries $\{Q_{t}\}$ are used to learn new appeared knowledge~(classes) $\{\mathcal{C}_{t}\}$; 
2) Knowledge distillation~\cite{(PLOP)douillard2020plop,zhang2022representation,douillard2020podnet} is a commonly used technique to overcome the challenge of catastrophic forgetting, where it typically involves directly distilling and aligning the features between $t$-th step and $(t-1)$-th to avoid a serious performance drop for old classes when learning new classes.
Different from previous knowledge distillation, query-guided knowledge distillation is more precise to transfer the knowledge from $f_{\bm\theta}^{t-1}$ to $f^t_{\bm\theta}$ at each class/instance level with query groups $\sum_{i=0}^{t-1} Q_i$, where the inter-class/instance diversity and intra-class/instance identity can be captured, as shown in Fig.~\ref{fig:3b}.
Specifically, existing knowledge distillation methods often directly extract features from the backbone and decoder for distillation. 
However, features at $t-1$ step  only contain characteristics of old classes,
when features at step $t-1$ contain characteristics of both old and new classes.
Forcing distillation on these two features can undermine the learning of new class features because features at step $t-1$ do not encompass the characteristics of new classes.

For $t$-th step, we use $\mathcal{D}_t$ = $\{\bm x_{t},\bm{y}_{t}\}$ to denote the current training set, where $\bm x_t\in \mathcal{X}$ denotes the input image, and the $\bm{y}_t\in \mathcal{Y}_t$ denotes the corresponding ground-truth (GT) \textit{pixel} labels.
The label space $\mathcal{Y}_t=\mathcal{B}_t\cup\mathcal{C}_t$ consists of the current newly added classes  $\mathcal{C}_t$ and the dummy background class $\mathcal{B}_t$. 
The $\mathcal{B}_t$ label may includes \textit{past} classes $\mathcal{C}_{0:t-1}$, the \textit{future} classes $\mathcal{C}_{t+1:T}$, or the true background pixels $\mathcal{\widetilde{B} }_t$. 
Therefore, the dummy background class will cause a serious performance drop, while the potential foreground classes~( $\mathcal{C}_{0:t-1}$, $\mathcal{C}_{t+1:T}$) is uncertain or misleading.
The challenge in continual image segmentation is named \textit{background shift}.
All previous works try to predict the potential foreground classes, \ie{} $\mathcal{C}_{0:t-1}$, $\mathcal{C}_{t+1:T}$ with the pseudo label from old model~\cite{(MiB)cermelli2020modeling,(PLOP)douillard2020plop} or other off-the-shelf saliency-map detector~\cite{cha2021ssul}. 
But the predicted pseudo-labeling from the model usually presents instability and worse performance, and it is impossible to predict all potential objects for future classes.
In our work, we introduce a new concept, ``\textit{Adaptive}'' background class, set $\mathcal{B}_t=\{c_t\}\cup\mathcal{\widetilde{B} }_t$, where $\{c_t\}$ refer to the potential foreground classes and is considered as background for training at $t$-th step.
$c_t$ is a variable and different for each class in a different step.
For instance, at step $t-1$, if only class \texttt{sofa} mask annotation is available,  \texttt{person}, \texttt{dog} pixel region as the \textit{unknown} regions belong to $\{c_{t-1}\}$.
But at step $t$, \texttt{person} as newly added class is defined as the positive sample, but \texttt{sofa} mask is not available, belonging to the `background' class $\{c_{t}\}$.
The adaptive background concept can completely \textit{solve} the negative impact from background shift, while we do not need to give a certain definition for what is real background in each step. 

\section{Approach}

\subsection{Dynamic Query}
Fig.~\ref{fig:3a} presents the whole framework for incremental Dynamic Query, which describes how to decouple the retaining of old knowledge and learning of new knowledge.
All previous CNN-based works~\cite{(PLOP)douillard2020plop,cha2021ssul} all adopt Deeplab v3~\cite{(deeplabv3)chen2017rethinking} as the base framework, which views segmentation task as the multi-classification per-pixel problem.
It is difficult to learn new classes and keep discrimination for old classes simultaneously for such a framework, where all representations are stored in the same embedding space.
Different from these works, we adopt query-based mask classification architecture, Mask2former~\cite{cheng2022masked}, which uses a set of $C$-dimensional feature vectors~(\ie{} ``query'') to predict a set of binary masks each associated with a single category.
Therefore, different representations from different categories or instances can be decoupled into different feature embedding with a different query.

\subsubsection{Incremental Dynamic Query Queue}
\label{DQ}
To enable continual learning in such architecture, we modify the fixed set of learned queries of Transformer decoder~\cite{cheng2022masked} to \textit{dynamic} incremental increasing sets. 
For the standard setting, given a input image $I$ and the base model $f_{\bm\theta}$ with parameter $\theta$, the predicted classes is denoted as  $\{\mathcal{C},\mathcal{B}\} = f_{\bm\theta}(Q|I)$, where $Q$ is a fixed set of queries, \ie{} $N$ $C$-dimensional learnable feature vectors.
$\mathcal{C}$ and $\mathcal{B}$ are the predicted foreground classes and background class.

The fixed set of queries is transformed to a dynamic incremental increasing query queue of $C$-dimensional vectors.
Specially, in step $0$, $Q_0$ include $n^q_0$ $C$-dimensional vectors, representing the base class set $\mathcal{C}_0$ with $n^c_0$ classes, where $n^q_0$ $>$ $n^c_0$.
And we model the current background class as $\mathcal{B}_0=\{c_0\}\cup\mathcal{\widetilde{B} }_0$, where $\widetilde{B}_0$ is the true background pixel, and $c_0$ is the potential foreground pixel defined as `background', \eg{} the region of \textit{past} class \texttt{person}.  
For the subsequent learning steps, the model is supposed to segment newly added classes $\{\mathcal{C}_1,\mathcal{C}_2,\mathcal{C}_3...\}$.
Corresponding, newly query sets $\{Q_1,Q_2,Q_3...\}$ will be added to guide the representation learning of newly added classes.
In the training phase of step $t$, query embedding sets $Q_{0:t-1}$ will be frozen to avoid forgetting old knowledge.
And $Q_t$ is used to guide the network to learn new knowledge of newly added classes.
Finally, we model the continual learning at $t$-th step as:
\begin{align}
\{\mathcal{C}_{0:t},\mathcal{B}_{0:t}\} = f_{\bm\theta}({\textcolor[RGB]{119,136,153}{Q_{0:t-1}} } ,Q_t|I)\,.
\end{align}
Actually, only $\{\mathcal{C}_{t},\mathcal{B}_{t}\}$ are supervised by manual annotation, while other annotations of $\{\mathcal{C}_{0:t-1},\mathcal{B}_{0:t-1}\}$ are not available.
Class/instance-aware query guided knowledge distillation~(Section~\ref{QKD}) is used to further retain the old knowledge of past classes.

\subsubsection{Adaptive Background for Different Query Set}
Section~\ref{DQ} have provided basic information concerning the adaptive backgrounds $\{\mathcal{B}_0,\mathcal{B}_1...\mathcal{B}_{T} \}$ for different step.
Actually, as shown in Fig.~\ref{fig:3a}, different from foreground classes $\mathcal{C}_{0:T}$, we do not use query to guide predict background mask, where a pixel belongs to background region $\mathcal{B}_{t}$ if it is not in foreground region $\mathcal{C}_{t}$ at step $t$.
During continual learning steps, the definition of background will change adaptively with the changing of foreground classes.
Therefore, the background shift problem can be solved naturally, while it is unnecessary to give a certain definition that which region is the true background, and no confused supervision.

\subsubsection{Independent Matching for Each Query Set}
Independent bipartite matching is designed for each query set at different learning step, which includes advantages: 1) Avoiding the mutual interference between old and new knowledge. 2) Enabling adaptive background.

Standard bipartite matching between ground truth $\bm{y}^i$ and a predictions $\hy^{\sigma(i)}$ with index $\sigma(i)$ computes:
\begin{equation}
\label{eq:matching}
    \hat{\sigma} = \argmin_{\sigma\in\Sigma_M} \sum_{i}^{M} \lmatch{\bm{y}^i, \hy^{\sigma(i)}},
\end{equation}
where $M$ is the number of predictions, the same as the number of queries.
$\lmatch{\bm{y}^i, \hy^{\sigma(i)}}$ is a pair-wise matching cost~\cite{carion2020end}.
For step $t$, we modify the matching to:
\begin{equation}
\label{eq:matching1}
    \hat{\sigma} = \argmin_{\sigma\in\Sigma_{M_t}} \sum_{i}^{M_t} \lmatch{\bm{y}^i_t, \hy^{\sigma(i)}_t},
\end{equation}
where $M_t$, $\bm{y}^i_t$, and $\hy^{\sigma(i)}_t$ refer to the number of newly added query, the ground truth of new classes, and the corresponding prediction. 
Therefore, only the prediction of the newly added classes is supervised with annotation for the network.
For the knowledge of old classes, we adopt \textit{freezing} query embedding and query guided knowledge distillation~(Sec.~\ref{QKD}) to retain them.

\begin{figure}
\centering
  \includegraphics[width=1.0\linewidth]{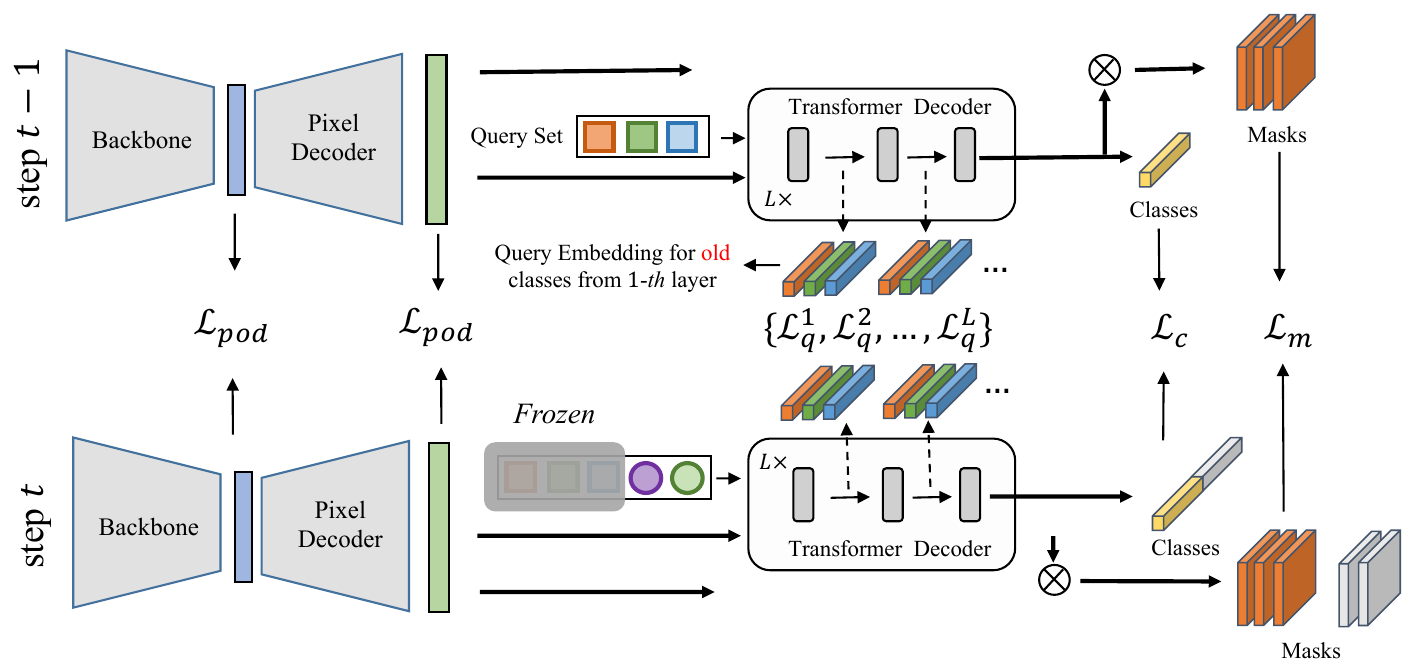}
  \caption{\textbf{Illustration of Query Guided Knowledge Distillation}. Query G-KD mainly includes three parts: 1) Multi-Scale KD with local POD~\cite{(PLOP)douillard2020plop,douillard2020podnet} $\mathcal{L}_{pod}$ for feature from backbone and pixel-decoder; 2) Class/Instance-Aware Query KD $\{\mathcal{L}_q^1,\mathcal{L}_q^2,\dots,\mathcal{L}_q^L\}$ for query embedding vectors from multi-layers of transformer decoder; 3) Class/Instance-Aware Prediction KD for network prediction, \ie{} class $\mathcal{L}_c$ and mask $\mathcal{L}_m$.}
  \vspace{-0.1cm}
\label{fig:KD}
\end{figure}

\subsection{Query Guided Knowledge Distillation}
\label{QKD}
As shown in Fig.~\ref{fig:3b}, existing knowledge distillation methods~\cite{(PLOP)douillard2020plop,zhang2022representation,douillard2020podnet} try to solve catastrophically forgetting via matching \textbf{global} statistics distribution at different feature levels between the old and current models.
Given an embedding tensor $\vx$, whose size is $H \times W \times C$.
These methods usually focus on how to extract a better representation of the embedding tensor with a mapping function $\Phi$~(\eg{}, PCKD~\cite{douillard2020podnet}, POD~\cite{zhang2022representation}),
then train the model $f^{t}_{\bm\theta}$ via minimizing the L2 distance between the two distribution:
\begin{equation}
    \mathcal{L}(\Theta^t) = \frac{1}{L} \sum_{l = 1}^L \left\Vert  \Phi(\vx^t_l) -  \Phi(\vx^{t-1}_l) \right\Vert^2\,,
\label{eq:pod_loss}
\end{equation}
where $L$ denotes the number of embedding tensors.
Actually, the \textbf{global} statistics match paradigm is crude and unreasonable with two drawbacks: 1) Directly matching will damage the representation learning of new classes, while $\vx^{t-1}_l$ from the previous model $f^{t-1}_{\bm\theta}$ do not contain the representation.
2) Inter-class diversity and intra-class identity can not be captured for indiscriminate matching.

To solve the two drawbacks, Class/Instance-Aware Query Guided Knowledge Distillation~(Query G-KD) is proposed.
Fig.~\ref{fig:KD} presents the whole illustration of Query G-KD, which includes three KD methods, \ie{} Multi-Scale KD with local POD, Class/Instance-Aware Query KD, and Class/Instance-Aware Prediction KD.

\subsubsection{Class/Instance-Aware Query KD}
At step $t$, given a $C$-dimensional learnable query embedding $\vq\in \{Q_0,Q_1,...,Q_{t-1}\}$ that is responsible for a single \textit{old} category $\mathcal{C}_{0:t-1}$, we can obtain $L$ query embedding vectors $\{\vq_1, \vq_2...\vq_{L}\}$ from the output of each layer at transformer decoder for all $L$ layers, as show in Fig.~\ref{fig:KD}.
In the training stage, The original query embedding $\vq$ is frozen and extracted query embedding vectors $\{\vq_1, \vq_2...\vq_{L}\}$ are used for distillation.
Then, the query knowledge distillation loss function $\mathcal{L}_q$ for the intermediate layers can be denoted as
\begin{equation}
    \mathcal{L}_q(\Theta^t) = \frac{1}{M} \frac{1}{L}  \sum_{j = 0}^M \sum_{l = 1}^L \left\Vert  \vq^{t}_{j,l} -  \vq^{t-1}_{j,l} \right\Vert^2\,,
\label{eq:local_pod_loss}
\end{equation}
where $M$ is the number of all query embeddings $Q_{0:t-1}$ for \textit{old} classes. And $\vq^{t}_{j,l}$ and $\vq^{t-1}_{j,l}$ refer to the student and teacher model from step $t$ and $t-1$, respectively.

\subsubsection{Class/Instance-Aware Prediction KD}
To further retain the knowledge of old classes, prediction knowledge distillation is proposed for match classification and mask prediction between the student and teacher models.
Given $t-1$ query groups $\{Q_0,Q_1,...,Q_{t-1}\}$ for old classes, we can obtain $t-1$ predicted class distribution vectors $\{\vc_0,\vc_1,...,\vc_{t-1}\}$ from the output of class head before Softmax activation function.  
Therefore, we can match the distribution of old classes between two models with Kullback-Leibler~(KL) divergence:
\begin{equation}
    \mathcal{L}_c(\Theta^t) = \sum_{i = 0}^{t-1} \vc_i^{t-1} \log \frac{\vc_i^{t-1}}{\vc_i^{t}} \,,
\label{eq:local_class}
\end{equation}
where $\vc_i^{t}$ and $\vc_i^{t-1}$ refer to $i$-th class distribution prediction associated with query set $Q_i\in\{Q_0,Q_1,...,Q_{t-1}\}$ from step $t-1$ model~(\ie{} teacher) and $t$ model~(\ie{} student), respectively.
Similarly, the knowledge distillation of the predicted mask can be computed with the binary cross-entropy loss~\cite{cheng2021per} $\mathcal{L}_{ce}$ and dice loss~\cite{milletari2016v} $\mathcal{L}_{dice}$:
\begin{equation}
\begin{aligned}
    \mathcal{L}_m(\Theta^t) &= \lambda_{c} \frac{1}{M} \sum_{j = 0}^{M} \mathcal{L}_{\text{ce}}(\vm_j^t,\vm_j^{t-1}) \\ 
    &+ \lambda_{d} \frac{1}{M} \sum_{j = 0}^{M} \mathcal{L}_{\text{dice}}(\vm_j^t,\vm_j^{t-1}) \,,
\end{aligned}
\label{eq:local_mask}
\end{equation}
where $M$ is the number of all query embeddings $Q_{0:t-1}$ for \textit{old} classes.
$\vm_j^{t}$ and $\vm_j^{t-1}$ denote the $j$-th predicted mask from step $t-1$ model~(\ie{} teacher) and $t$ model~(\ie{} student), respectively.
$\lambda_{c}$ and $\lambda_{d}$ are two weight parameters, similar to Mask2former~\cite{cheng2022masked}.

\begin{algorithm}[t]
    \small
    \caption{Training pipeline of CisDQ}
    \label{alg:cisdq}
    \begin{algorithmic}[1]
        \Require {$\mathcal{C}_0, \mathcal{I}_0$: the base class set and corresponding image set.}
        \Require {$\{(\mathcal{C}_1, \mathcal{I}_1), (\mathcal{C}_2, \mathcal{I}_2), \cdots, (\mathcal{C}_T, \mathcal{I}_T)\}$ : the novel class set and corresponding image set for each incremental step.}
        \Require {$\alpha$ : the learning rate.}
        \State {Initialize model parameters $\bm\theta$ and the base query set $Q_0$ for CisDQ model $f$}
        \State {Optimize $\{\bm\theta, Q_0\}$ with the Mask2former loss on image set $\mathcal{I}_0$}
        \For {incremental step $t \in \{1, \cdots, T\}$}
            \State {$\triangleright$ $Q = \{Q_i\}_{i=0}^{t-1}\cup Q_t$, Update the query set with random} 
            \State {$\quad$initialized novel query}
            \State {$\triangleright$ Freeze $\{Q_i\}_{i=0}^{t-1}$}
            \For {\textbf{each} training step}
                \State $\triangleright$ Sample a mini-batch of training images $I$, $I \in \mathcal{I}_t$
                \State $\triangleright$ $\hy = f_{\bm\theta}(Q_{0:t}|I)$, get network predictions $\hy$
                \State $\triangleright$ Match the predictions $\hy$ to the ground truth $\bm{y} \in \mathcal{C}_t$
                \State $\quad$with Equ.~\ref{eq:matching1}
                \State $\triangleright$ Calculate loss $\mathcal{L}$ with Equ.~\ref{eq:complete_loss}:
                \State $\quad\quad \mathcal{L} = \strut \mathcal{L}_\text{new} + \strut \lambda_1\mathcal{L}_q+\lambda_2\mathcal{L}_c+\lambda_3\mathcal{L}_m+\lambda_4\mathcal{L}_{pod}$
                \State $\triangleright$ Update $\{\bm\theta, Q_t\}$ with back propagation algorithm:
                \State $\quad\quad \{\bm\theta, Q_t\} = \{\bm\theta, Q_t\} - \alpha\nabla \mathcal{L}$
            \EndFor
        \EndFor
        \Ensure{The trained CisDQ model $f$ with weights $\{\bm\theta, Q_{0:T}\}$}
    \end{algorithmic}
\end{algorithm}

\begin{table}[h]
    \centering
    \small 
    \setlength{\tabcolsep}{1mm}
    

\centering
\setlength{\tabcolsep}{1.3mm}
\footnotesize
\begin{tabular}{ l|c|c} 

Settings& Step Number & Few Shot for New Class \\ 
\hline
iFSIS & Two-Step~(Step 0, 1) & \checkmark \\
CIS &  Multi-Step~(Step 0, 1, \dots, n) &  \\

\end{tabular}

    \caption{\textbf{Comparison of Related Task.} `iFSIS' and `CIS' refer to incremental few-shot instance segmentation and continual instance segmentation, respectively.}
    \label{Task}
\end{table}

\begin{table}[h]
    \centering
    \normalsize 
    \setlength{\tabcolsep}{1mm}
    

\centering
\setlength{\tabcolsep}{0.4mm}
\tiny
\begin{tabular}{ l|l} 

Seq.& 80 Instance Classes on COCO 2017  \\ 
\hline
\textbf{1-8} & person, bicycle, car, motorcycle, airplane, bus, train, truck  \\
\textbf{9-16} & boat, traffic light, fire hydrant, stop sign, parking meter, bench, bird, cat \\
\textbf{17-24} & dog, horse, sheep, cow, elephant, bear, zebra, giraffe\\
\textbf{25-32} & backpack, umbrella, handbag, tie, suitcase, frisbee, skis, snowboard\\
\textbf{33-40} &  sports ball, kite, baseball bat, baseball glove, skateboard, surfboard, tennis racket, bottle\\
\textbf{41-48} &
wine glass, cup, fork, knife, spoon, bowl, banana, apple\\
\textbf{49-56} &
sandwich, orange, broccoli, carrot, hot dog, pizza, donut, cake\\
\textbf{57-64} &
chair, couch, potted plant, bed, dining table,toilet, tv, laptop\\
\textbf{65-72} & mouse, remote, keyboard, cell phone, microwave, oven, toaster, sink\\
\textbf{72-80} & refrigerator, book, clock, vase, scissors, teddy bear, hair drier, toothbrush\\

\end{tabular}

    \caption{\textbf{Class Order, for Instance, Segmentation with 80 Classes on COCO 2017 .}}
    \label{cocooder}
\end{table}

\begin{table}[h]
    \centering
    \normalsize 
    \setlength{\tabcolsep}{1mm}
    

\centering
\setlength{\tabcolsep}{0.4mm}
\tiny
\begin{tabular}{ l|l} 

Seq.& 100 Instance Classes on ADE 20k  \\ 
\hline
\textbf{1-10} & bed, windowpane, cabinet, person, door, table, curtain, chair, car, painting,  \\
\textbf{11-20} & sofa, shelf, mirror, armchair, seat, fence, desk, wardrobe, lamp, bathtub, \\
\textbf{21-30} & railing, cushion, box, column, signboard, chest of drawers, counter, sink, fireplace, refrigerator, \\
\textbf{31-40} & stairs, case, pool table, pillow, screen door, bookcase, coffee table, toilet, flower, book ,\\
\textbf{41-50} & bench, countertop, stove, palm, kitchen island, computer, swivel chair, boat, arcade machine, bus, \\
\textbf{51-60} & towel, light, truck, chandelier, awning, streetlight, booth, television receiver, airplane, apparel, \\
\textbf{61-70} & pole, bannister, ottoman, bottle, van, ship, fountain, washer, plaything, stool, \\
\textbf{71-80} & barrel, basket, bag, minibike, oven, ball, food, step, trade name, microwave, \\
\textbf{81-90} & pot, animal, bicycle, dishwasher, screen, sculpture, hood, sconce, vase, traffic light, \\
\textbf{91-100} & tray, ashcan, fan, plate, monitor, bulletin board, radiator, glass, clock, flag \\
\end{tabular}

    \caption{\textbf{Class Order, for Instance, Segmentation with 100 Classes on ADE 20k.}}
    \label{adeorde}
\end{table}

\begin{table*}[t]
    \centering
    \small 
    \setlength{\tabcolsep}{1mm}

\centering
\setlength{\tabcolsep}{1.3mm}
\footnotesize
\begin{tabular}{ l||c||c||cc|c||cc|c||cc|c||cc|c } 
\toprule
\multirow{2}*{\textbf{Method}}&
\multirow{2}*{\textbf{Base Network}}&
\multirow{2}*{\textbf{Backbone}} & \multicolumn{3}{c||}{\textbf{ADE 100-50 (2 steps)}} & \multicolumn{3}{c||}{\textbf{ADE 100-10 (6 steps)}} & \multicolumn{3}{c||}{\textbf{ADE 50-50 (3 steps)}} & \multicolumn{3}{c}{\textbf{ADE 100-5 (11 steps)}}\\
 &  & & \textit{1-100} & \textit{101-150} & \textit{all} & \textit{1-100} & \textit{101-150} & \textit{all} & \textit{1-50} & \textit{51-150} & \textit{all} & \textit{1-100} & \textit{101-150} & \textit{all} \\ 
\midrule
\demph{Joint Training} & \demph{Deeplab V3} & \demph{ResNet-101} & \demph{44.3} & \demph{28.2} & \demph{38.9} & \demph{44.3} & \demph{28.2} & \demph{38.9} & \demph{51.1} & \demph{33.3} & \demph{38.9} & \demph{44.3} & \demph{28.2} & \demph{38.9}\\
ILT~\cite{(ILT)michieli2019incremental} & Deeplab V3 & ResNet-101 & 18.3 & 14.4 & 17.0 & 0.1 & 3.1 & 1.1 & 3.5 & 12.9 & 8.7 & 0.1 & 1.3 & 0.5 \\
MiB~\cite{(MiB)cermelli2020modeling} & Deeplab V3 & ResNet-101 & 40.5 & 17.2 & 32.8 & 38.2 & 11.1 & 29.2 & 45.6 & 21.0 & 29.3 & 36.0 & 5.7 & 26.0 \\
PLOP~\cite{(PLOP)douillard2020plop} & Deeplab V3 & ResNet-101 & 41.9 & 14.9 & 32.9 & 40.5 & 13.6 & 31.6 & 48.8 & 21.0 & 30.4 & 39.1 & 7.8 & 28.8 \\ 
RCIL~\cite{zhang2022representation} & Deeplab V3 & ResNet-101 & 42.3 & 18.8 & 34.5 & 39.3 & 17.6 & 32.1 & 48.3 & 25.0 & 32.5 & 38.5 & 11.5 & 29.6 \\ 
MiB+EWF~\cite{xiao2023endpoints} & Deeplab V3 & ResNet-101 & 41.2 & 21.3 & 34.6 & 41.5 & - & 33.2 & -& - & - & 41.4 & 13.4 & 32.1
\\
\demph{SSUL~\cite{cha2021ssul}} & \demph{Deeplab V3} & \demph{ResNet-101} & \demph{41.3} & \demph{18.0} & \demph{33.6} & \demph{40.2} & \demph{18.8} & \demph{33.1} & \demph{48.4} & \demph{20.2} & \demph{29.6} & \demph{39.9} & \demph{17.4} & \demph{32.5}\\
\midrule
\demph{Joint Training~$^{\star}$} & \demph{Mask2former}  & \demph{ResNet-50} & \demph{49.8} & \demph{36.9} & \demph{45.5} & \demph{49.8} & \demph{36.9} & \demph{45.5} & \demph{55.9}
& \demph{40.3} & \demph{45.5} & \demph{49.8} & \demph{36.9} & \demph{45.5} \\

CoMFormer~\cite{cermelli2023comformer} & Mask2former & ResNet-101 & 44.7 & 26.2 & 38.4 & 40.6 & 15.6 & 32.3 & - & -& - & 39.5 & 13.6 & 30.9 \\
MiB~$^{\star}$ & Mask2former  & ResNet-50 & 47.6
& 27.8 & 41.0 & 40.6 & 16.3 & 32.5 & 52.7 & 30.8 & 38.1 & 38.0 & 12.2 & 29.4 \\ 
PLOP~$^{\star}$ & Mask2former  & ResNet-50 & 48.2
& 28.0 & 41.5 & 44.3 & 18.6 & 35.7 & 55.5 & 32.0 & 39.9 & 39.2 & 13.2 & 30.5 \\ 
MicroSeg~\cite{zhang2022mining} & Mask2former & ResNet-101 & 40.2 & 18.8 & 33.1 & 41.5 & 21.6 & 34.9 & 48.6 & 24.8 & 32.9 & 40.4 & 20.5& 33.8  \\ 

MicroSeg-M~\cite{zhang2022mining} & Mask2former & ResNet-101 & 43.4 & 20.9 & 35.9 & 43.7 & 22.2 & 36.6 & 49.8 & 22.0 & 31.4 & 43.6& 22.4& 36.6\\
\textbf{CiSDQ} (ours) & Mask2former  & ResNet-50 & \textbf{48.9} & \textbf{28.2} & \textbf{42.0} & \textbf{47.8} & \textbf{24.6} & \textbf{40.1} & \textbf{55.7} & \textbf{33.6} & \textbf{41.0} & \textbf{46.2} & \textbf{17.6} & \textbf{36.7}\\
\bottomrule
\end{tabular}

    \caption{\textbf{The final mIoU(\%) of Semantic Segmentation on the ADE20K dataset.} $^{\star}$ refers to our re-implementation performance. \demph{In gray} denotes joint training with all classes or using extra model~(\ie{} SSUL using DSS~\cite{hou2017deeply} pretrained on MSRA-B dataset~\cite{liu2010learning}).}
    \label{ADE20k_semantic}
\end{table*}

\begin{table*}[t]
    \centering
    \small 
    \setlength{\tabcolsep}{1mm}

\centering
\setlength{\tabcolsep}{1.8mm}
\footnotesize
\begin{tabular}{ l||c||c||cc|c||cc|c||cc|c } 
\toprule
\multirow{2}*{\textbf{Method}}&\multirow{2}*{\textbf{Base Network}} & \multirow{2}*{\textbf{Backbone}} & \multicolumn{3}{c||}{\textbf{VOC 15-1 (6 steps)}} & \multicolumn{3}{c||}{\textbf{VOC 15-5 (2 steps)}} & \multicolumn{3}{c}{\textbf{VOC 10-1 (11 steps)}} \\
 &  & & \textit{0-15} & \textit{16-20} & \textit{all} & \textit{0-15} & \textit{16-20} & \textit{all} & \textit{0-10} & \textit{11-20} & \textit{all} \\ 
\midrule
\demph{Joint Training~(offline)} & \demph{Deeplab V3} & \demph{ResNet-101} & \demph{79.8} & \demph{72.3} & \demph{77.4} & \demph{79.8} & \demph{72.4} & \demph{77.4}  & \demph{78.4} & \demph{76.4} & \demph{77.4}\\
LwF-MC~\cite{(LwF)LiHoiem16} & Deeplab V3 & ResNet-101 & 6.4 & 8.4 & 6.9 & 58.1 & 35.0 & 52.3 & 4.7 & 5.9 & 5.0 \\
ILT~\cite{(ILT)michieli2019incremental} & Deeplab V3 & ResNet-101 & 8.8 & 8.0 & 8.6 & 67.1 & 39.2 & 60.5  & 7.2 & 3.7 & 5.5 \\
MiB~\cite{(MiB)cermelli2020modeling} & Deeplab V3 & ResNet-101 & 35.1 & 13.5 & 29.7 & 75.5 & 49.4 & 69.0  & 12.2 & 13.1 & 12.6 \\
PLOP~\cite{(PLOP)douillard2020plop} & Deeplab V3 & ResNet-101 & 65.1 & 21.1 & 54.6 & 75.7 & 51.7 & 70.1  & 44.0 & 15.5 & 30.5 \\ 
RCIL~\cite{zhang2022representation} & Deeplab V3 & ResNet-101 & 70.6 & \textbf{23.7} & 59.4 & 78.8 & \textbf{52.0} & 72.4 & 55.4 & 15.1 & 34.3 \\ 
\demph{SSUL~\cite{cha2021ssul}} & \demph{Deeplab V3} & \demph{ResNet-101} & \demph{77.3} & \demph{36.6} & \demph{67.6} & \demph{77.8} & \demph{50.1} & \demph{71.2}  & \demph{71.3} & \demph{46.0} & \demph{59.3}\\

\midrule
\demph{Joint Training~(offline)~$^{\star}$} & \demph{Mask2former}  & \demph{ResNet-50} & \demph{80.0} & \demph{75.2} & \demph{78.8} & \demph{80.0} & \demph{75.2} & \demph{78.8}  & \demph{78.9} & \demph{78.8} & \demph{78.8} \\
PLOP~$^{\star}$ & Mask2former  & ResNet-50 & 71.3 & 12.5 & 57.3 & 78.2 & 28.4 & 66.3 & 56.9 & 14.5 & 36.7 \\ 
\textbf{CiSDQ} (ours) & Mask2former  & ResNet-50 & \textbf{77.9} & 13.2 & \textbf{62.5} & \textbf{80.5} & 49.3 & \textbf{73.1} &  \textbf{73.2} & \textbf{15.5} & \textbf{45.7} \\
\textbf{CiSDQ} (ours) & Mask2former  & ResNet-101 & 78.2 & 13.6 & 62.8 & 80.6 & 49.5 & 73.2 &  73.9 & 15.8 & 46.2 \\
\textbf{CiSDQ} (ours) & Mask2former  & Swin-B & 79.7 & 14.9 & 64.3 & 84.0 & 55.1 & 77.1&  74.6 & 15.9 & 46.7 \\
\bottomrule
\end{tabular}

    \caption{\textbf{The mIoU(\%) of Semantic Segmentation on the Pascal VOC 2012 dataset.} $^{\star}$ refer to our re-implementation performance. \demph{In gray} denotes joint training with all classes or using extra model~(\ie{} SSUL using DSS~\cite{hou2017deeply} pretrained on MSRA-B dataset~\cite{liu2010learning}).}
    
    \label{VOC_semantic}
\end{table*}

\subsection{Loss Function}
The proposed pipeline mainly contains two losses, \ie{} manual annotation supervision loss, and knowledge distillation loss.
The whole loss function can be formulated as Equation~\ref{eq:complete_loss}:
\begin{equation}
    \mathcal{L} = \underbrace{\strut \mathcal{L}_\text{new}}_\text{New Classes} + \underbrace{\strut \lambda_1\mathcal{L}_q+\lambda_2\mathcal{L}_c+\lambda_3\mathcal{L}_m+\lambda_4\mathcal{L}_{pod}\,\,}_\text{Distillation for Old Classes} \!\! \,,
\label{eq:complete_loss}
\end{equation}
where $\mathcal{L}_\text{new}$ denotes the supervision from the available annotation of newly added classes, the same as that of Mask2former~\cite{cheng2022masked}.
$\lambda_1$, $\lambda_2$, $\lambda_3$, $\lambda_4$ are the weight parameters, which are set to $1$, $5$, $300$, $100$, respectively.
$\mathcal{L}_{pod}$ is the local POD loss~\cite{douillard2020podnet}, the same as PLOP~\cite{(PLOP)douillard2020plop}, which is used to distill the features from backbone and pixel decoders.

\subsection{Pseudo Code}
We describe the training pipeline of CisDQ in Algorithm~\ref{alg:cisdq}. 
We first train the model on the image set of the base classes (line 1-2).
Then, for each incremental step, we add a set of new queries for the novel classes to the model and freeze the queries of the old classes (line 4-6).
After that, we optimize the model with the proposed matching strategy (line 10) and loss function (line 12-15). Finally, we get the CisDQ model that is capable to segment objects of all classes.

\section{Experiments}

\subsection{Experimental Setups}

\subsubsection{Semantic Segmentation}
\textbf{Datasets.} Similar to previous works~\cite{(PLOP)douillard2020plop,cha2021ssul}, Pascal-VOC 2012~\cite{(voc)everingham2010pascal} (20 classes) and ADE20k~\cite{(ade)zhou2017scene} (150 classes) are used to evaluate our \cisdq.
\textbf{Protocols.} Following works~\cite{(PLOP)douillard2020plop,cha2021ssul,cermelli2020modelingthebackground}, the model is trained to segment new classes in multiple steps continually.
And only new classes in the current step are labeled.
And there are two different CSS settings: \textit{Disjoint} and \textit{Overlapped}. 
For step $t$, images of \textit{Disjoint} only contain classes $\mathcal{C}^{1:t-1} \cup \mathcal{C}^{t}$ (old and new), while that of \textit{Overlapped} can includes any classes $\mathcal{C}^{1:t-1} \cup \mathcal{C}^{t} \cup \mathcal{C}^{t+1:T}$ (old, new, and future). 
Therefore, the Overlapped setting is more challenging and realistic.
%
In our experiments, we only focus on the result concerning Overlapped CSS.
During the testing phase, the model need to segment all classes.
Protocol setting following previous works~\cite{(PLOP)douillard2020plop,zhang2022representation}, \eg{} ADE 100-50, 100-10, 50-50, and 100-5, which consists in learning $100$ classes followed by 50 class ($2$ steps), $100$ classes followed by five times $10$ classes ($6$ steps), $50$ classes followed by two times $50$ classes ($3$ steps), and $100$ classes followed by ten times $10$ classes ($11$ steps).
%
%

\begin{figure}[t]
	\includegraphics[width=0.99\linewidth]{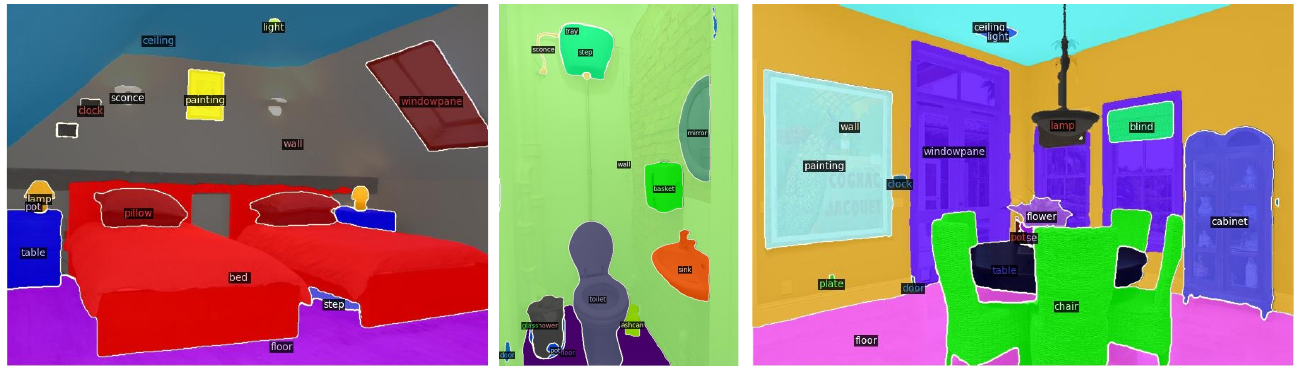}
	\caption{\textbf{More visualization of \cisdq for semantic segmentation on ADE 20k \texttt{val}.} \cisdq presents high-quality results.}
\label{VIS12}
\end{figure}
\subsubsection{Instance Segmentation}
\textbf{Datasets.} Following general instance segmentation~\cite{cheng2022masked,he2017mask}, COCO~\cite{lin2014microsoft} (80 classes) and ADE20k~\cite{(ade)zhou2017scene} (100 classes) are used to evaluate our \cisdq.
\textbf{Protocols.} Different from CCS, Continual Instance Segmentation~(CIS), as one new task, need the first definition for experiment settings. 
As for COCO~\cite{lin2014microsoft}, we give three-step settings, \ie{} 40-40 (2 steps), 40-8 (6 steps), and 40-4 (11 steps).
For ADE20k, similar to the class split of CCS, we provide three step settings concerning two steps (50-50), six steps (50-10), and eleven steps (50-5).
The detailed classes split for each step and related experiments for ADE20k are provided in in supplementary material.
%
%
Similar to CCS, we focus on \textit{Overlapping} setting, which is more realistic and challenging.

\subsubsection{Implementation Details.} 
The most experiment setting of the experiments all follow Mask2former~\cite{cheng2022masked}, \ie{} backbone, transformer, pixel decoder, AdamW optimizer with an initial learning rate of 0.0001.
And 8 Tesla V100 GPUs are used for the experiment.
For \textit{semantic} segmentation, the number of the query is the same as the that of class during continual learning, \eg{} ADE 100-50~(2 steps) require 100 queries to learn 1-100 classes at step 0, then increasing 50 queries to learn 101-150 classes at step 1.  
For \textit{instance} segmentation, the number of the query is set to $2.5$ times of that of class during continual learning, COCO requires 200 queries no matter which settings.
COCO 40-20~(3 steps) requires 100, 50, 50 queries for $0$-th, $1$-th, $2$-th steps, respectively.

\subsubsection{Compared with iFSIS.} Tab.~\ref{Task} compares continual instance segmentation~(CIS)  with other related problem, \ie{} incremental few-shot instance segmentation~\cite{ganea2021incremental,nguyen2022ifs}.
Different from continual segmentation, iFSIS only incremental learning on base and novel classes with just two steps, and the training data for new classes is limited and few.

\subsubsection{Protocols for Continual Instance Segmentation Datasets.} \textit{COCO 2017.} Different from continual semantic segmentation, Continual Instance Segmentation~(CIS), as one new task, needs the first definition for experiment settings. 
As for COCO 2017~\cite{lin2014microsoft}, we give three-step settings, \ie{} 40-40 (2 steps), 40-8 (6 steps), and 40-4 (11 steps).
And Table.~\ref{cocooder} provides the corresponding class order.
%
%
For ADE20k, similar to the class split of CCS, we provide three-step settings concerning two steps (50-50), six steps (50-10), and eleven steps (50-5).
The detailed class order for splitting is provided in Table.~\ref{adeorde}.
Similar to CCS, all experiments concerning instance segmentation task focus on \textit{Overlapping} setting, which is more realistic and challenging.

\begin{table*}[t]
    \centering
    \small 
    \setlength{\tabcolsep}{1mm}

\centering
\setlength{\tabcolsep}{1.8mm}
\footnotesize
\begin{tabular}{ l||cc|c||cccccc|c||cc|c} 
\toprule
\multirow{2}*{\textbf{Method}}&
\multicolumn{3}{c||}{\textbf{COCO 40-40 (2 steps)}} &
\multicolumn{7}{c||}{\textbf{COCO 40-8 (6 steps)}} & \multicolumn{3}{c}{\textbf{COCO 40-4 (11 steps)}} \\
 &  \textit{1-40} & \textit{41-80} & \textit{mAP}  & 
 \textit{1-40} & \textit{41-48} & \textit{49-56} & \textit{57-64} & \textit{65-72} & \textit{73-80} & \textit{mAP} & \textit{1-40} & \textit{41-80} &  \textit{mAP}  \\   
\midrule
\demph{Joint Training~(offline)} & \demph{35.6} & \demph{51.8} & \demph{43.7} & \demph{35.6} & \demph{47.0} & \demph{59.2} & \demph{53.8} & \demph{52.6} & \demph{46.2} & \demph{43.7} & \demph{35.6} & \demph{51.8} & \demph{43.7}\\
PLOP $^*$ & 28.0 & 40.1 & 34.1 & \textbf{24.4}  & 16.4 & 28.0 & 22.3 & 25.1 & 16.3 & 23.0 & 16.5 & 11.3 & 13.9\\
\textbf{CiSDQ} &  \textbf{28.8} & \textbf{41.8} & \textbf{35.3} & 24.2 & \textbf{17.9} & \textbf{29.4} & \textbf{24.8 }& \textbf{27.6} & \textbf{19.8} & \textbf{24.1} & \textbf{18.2} & \textbf{11.7} & \textbf{15.0} \\

\bottomrule
\end{tabular}

    \caption{\textbf{The final AP(\%) of Instance Segmentation on COCO \textit{val2017} with 80 categories.} Mask2former with ResNet50 is used as the base network. `offline' refers to training with all data and classes. $^{\star}$ refers to our re-implementation performance.}
    
    \label{ADE20k_COCO}
  \end{table*}

\begin{table*}[t]
    \centering
    \small 
    \setlength{\tabcolsep}{3mm}

\centering
\setlength{\tabcolsep}{1.8mm}
\footnotesize
\begin{tabular}{ l||cc|c||cccccc|c||cc|c } 
\toprule
\multirow{2}*{\textbf{Method}}&
%
\multicolumn{3}{c||}{\textbf{ADE 50-50 (2 steps)}} & \multicolumn{7}{c||}{\textbf{ADE 50-10 (6 steps)}} & \multicolumn{3}{c}{\textbf{ADE 50-5 (11 steps)}}  \\
 &   \textit{1-50} & \textit{51-100} & \textit{all} & \textit{1-50} & \textit{51-60} & \textit{61-70}& \textit{71-80} & \textit{81-90} & \textit{91-100} & \textit{all} & \textit{1-50} & \textit{51-100} & \textit{all}  \\ 
\midrule
\demph{Joint Training~(offline)} & \demph{32.2} & \demph{20.6} & \demph{26.4} & \demph{32.2} & \demph{24.6} & \demph{14.2} & \demph{16.9} & \demph{26.7} & \demph{20.5} & \demph{26.4} & \demph{32.2} & \demph{20.6} & \demph{26.4} \\
PLOP~$^{\star}$ & 23.4 & 11.8 & 17.6 & 13.1 & 6.4 & 2.1 & 7.2 & 11.1 & 4.9 & 9.7 & 12.9 & 2.5 & 7.7  \\ 
\textbf{CiSDQ} (ours) & 31.2 & 13.8 & 23.7 & 26.2 & 7.1 & 3.9 & 7.9 & 9.8 & 5.0 & 16.4 & 18.1 & 2.9 & 10.5 \\

\bottomrule
\end{tabular}

    \caption{\textbf{The final AP(\%) of Instance Segmentation on the ADE20K dataset.} $^{\star}$ refers to our re-implementation performance. ResNet50 as the backbone is used.}
    
    \label{ADE20k_instance}
\end{table*}

\subsection{Continual Semantic Segmentation}
\textbf{PASCAL VOC 2012.} Table.~\ref{VOC_semantic} presents the experimental results of the last step for three continual learning settings.
Compared with previous methods based on deeplab v3, our \cisdq achieves obvious mIoU improvements, especially for long-term steps setting, \eg{} VOC 10-1 (11 steps) and 15-1 (6 steps).
For a fair comparison, the performance of SSUL~\cite{cha2021ssul} is not the main reference, while it uses the extra Salient Object Segmentation model pre-trained on MSRA-B dataset~\cite{liu2010learning}).
Actually, the Salient object segmentation model usually can segment and obtain competitive pseudo label effectively, especially for VOC 2012, where the image domain is simple and only support 20 classes.
And we also re-implementation PLOP~\cite{(PLOP)douillard2020plop} with Mask2former framework.
We do not re-implementation RCIL~\cite{zhang2022representation} and SSUL~\cite{cha2021ssul} on Mask2former architecture due to the mismatched re-parameterization on transformer architecture from RCIL and using extract model of SSUL.
Overall, our \cisdq achieves state-of-the-art performance with obvious improvements no matter for the old or new classes.
Besides, we also provide more high-quality visualization results for semantic segmentation tasks on ADE 20k, as shown in Fig.~\ref{VIS12}.

\textbf{ADE20k.} Table.~\ref{ADE20k_semantic} presents the experimental results of the last step for four continual learning settings on ADE 20K.
On different continual learning tasks, \ie{} 100-50, 100-10 and 50-50, 100-5, our method all achieves state-of-the-art results, especially for the long step tasks, \ie{} 100-10~(6 steps) and 100-5~(11 steps), achieving $4.4\%$ and $2.9\%$ mIoU improvement over that of previous SOTA method, respectively.
Our algorithm demonstrates significant improvements compared to the ten methods based on Deeplab v3 or Mask2Former, substantiating the effectiveness of our approach.
For fair comparison, the performance of SSUL~\cite{cha2021ssul} and MicroSeg-M~\cite{zhang2022mining} is not the main reference, while it uses the extra Salient Object Segmentation model pre-trained on MSRA-B dataset~\cite{liu2010learning}).
And we also re-implementation MiB~\cite{(MiB)cermelli2020modeling} and PLOP~\cite{(PLOP)douillard2020plop} with Mask2former framework.
We do not re-implementation RCIL~\cite{zhang2022representation} and SSUL~\cite{cha2021ssul} on Mask2former due to the mismatched re-parameterization on transformer architecture of RCIL and the extra model of SSUL.
%

\begin{figure*}[t]
	\includegraphics[width=0.99\linewidth]{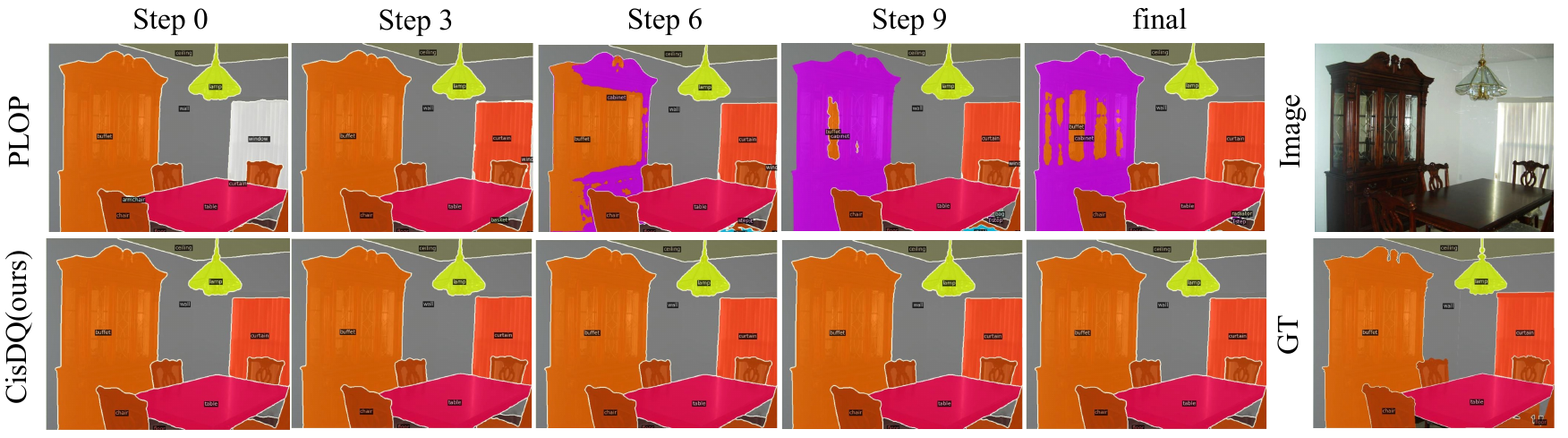}
	\caption{\textbf{Visualization of PLOP and \cisdq for \textit{100-5} (11 steps) on ADE 20k \texttt{val}.} \cisdq presents more robust performance for remaining the knowledge of old classes~( \eg{} \texttt{buffet}), while PLOP~\cite{(PLOP)douillard2020plop} suffer from the catastrophic forgetting.}
\label{VIS}
\end{figure*}

\begin{figure}[t]
	\includegraphics[width=0.99\linewidth]{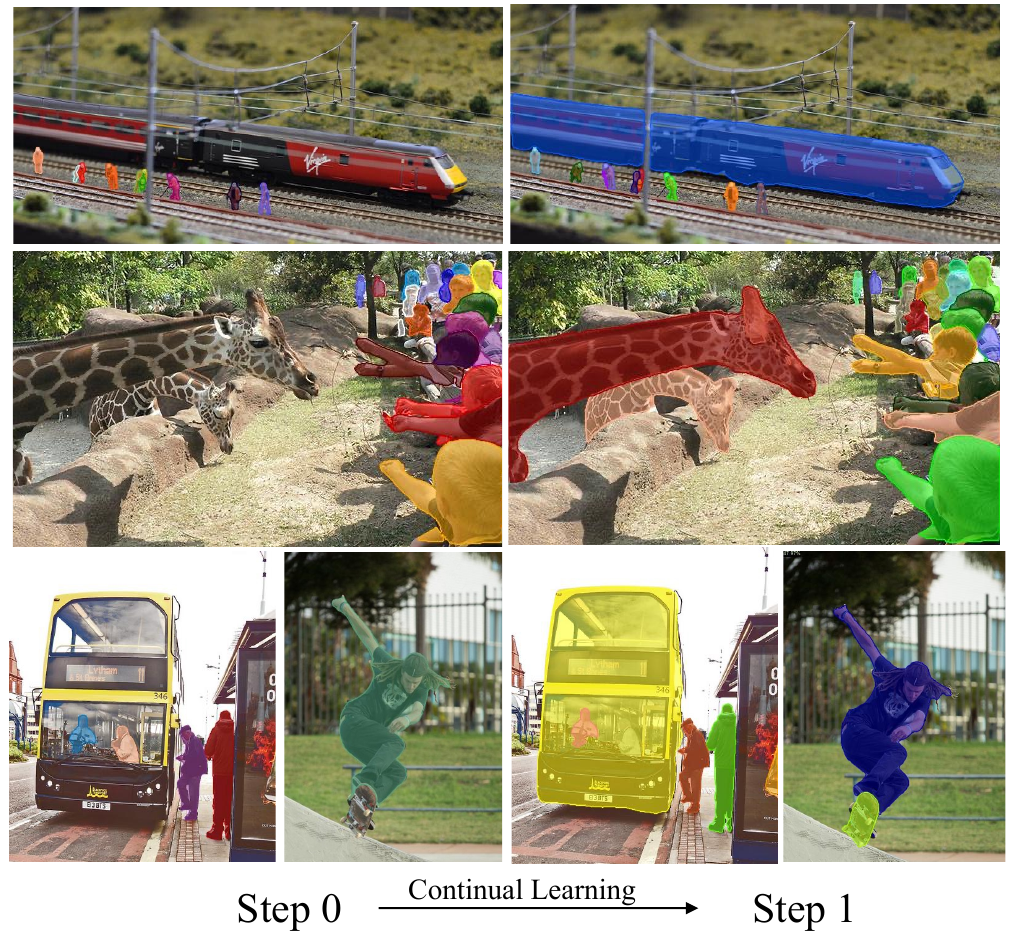}
	\caption{\textbf{Visualization of \cisdq for continual instance segmentation on COCO 2017 \texttt{val}.} \cisdq presents robust performance on learning new classes and remaining old knowledge.}
\label{VIS1}
\end{figure}

\subsection{Continual Instance Segmentation}
\textbf{COCO2017.} To verify the effectiveness of \cisdq for instance segmentation, we provide corresponding experiments on COCO~\cite{lin2014microsoft} for three different continual tasks, \ie{} COCO 40-40, 40-8, and 40-4, as shown in Table.~\ref{ADE20k_COCO}.
\cisdq presents competitive performance, around $1.0\%$ mAP improvement over previous methods.
It is worth noting that \cisdq presents a better performance for newly added classes, power from more reasonable KD for class/instance level.
%

\textbf{ADE 20k.}
We also provide the results for continual instance segmentation with three continual settings on ADE 20k, as shown in Fig.~\ref{ADE20k_instance}.
We implement PLOP~\cite{(PLOP)douillard2020plop} on Mask2former architecture for continual instance segmentation, while there is no one continual instance segmentation method for comparison.
\cisdq achieves obvious improvements over PLOP, with $6.1\%$, $6.7\%$, and $2.8\%$ mAP for the three different settings, respectively.
Besides, we also provide some visualizations for continual instance segmentation on COCO 2017, as shown in Fig.~\ref{VIS1}.

\begin{figure}[t]
	\includegraphics[width=0.90\linewidth]{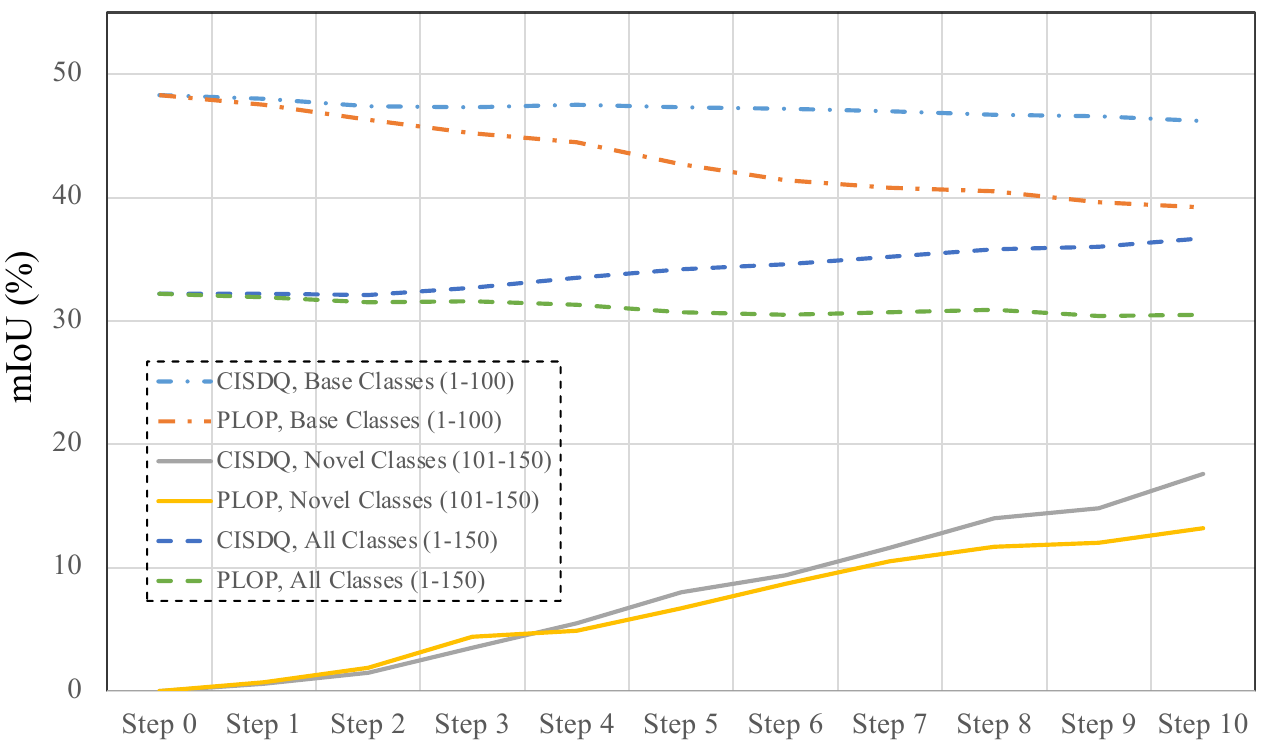}
	\caption{\textbf{Compared with PLOP~\cite{(PLOP)douillard2020plop} for semantic segmentation task on ADE 20k 100-5 (11 steps).} \cisdq presents better performance.}
\label{VIS123}
\end{figure}

\begin{table*}[t]
  
  \begin{subtable}{0.48\linewidth}
  \centering
  \tablestyle{2.6pt}{1.1}
  \scriptsize
  \begin{tabular}{cc|ccc}
   Incremental Query & Independent Match & 1-100 & 101-150 & \textit{all} \\
  \shline
   &  & 39.2 & 13.2 & 30.5 \\
   \checkmark &  & 39.6 \dt{+0.4} & 15.3 \dt{+2.1} &  31.5\dt{+1.0} \\
   \checkmark & \checkmark & 46.2 \dt{+6.6} & 17.6 \dt{+2.3} & 36.7 \dt{+5.2}\\
  \end{tabular}
  \caption{\textbf{Dynamic Query Queue}. Experiments are conducted on \textit{100-5} (11 steps) setting for semantic segmentation~(mAP) task on ADE20k \texttt{val}.
  }
  \label{ablation_DQ1}
  \end{subtable}\hspace{2mm}
  \begin{subtable}{0.48\linewidth}
  \centering
  \tablestyle{2pt}{1.2}
  \scriptsize
  \begin{tabular}{lcc|ccc|c}
  Method & Base Network & Backbone & Params & Flops & FPS & \textit{mIoU}\\
  \shline
  SSUL & Deeplab V3 & ResNet-101 & 62.7 & 255.1 & 14.1 & 33.1 \\
  CiSDQ & Mask2former & ResNet50 & 44.0 & 71.0 & 9.7 & 40.1 \\
  \end{tabular}
  \caption{\textbf{Base Network.} Experiments are conducted on \textit{100-10} setting for semantic segmentation~(mIoU) task on ADE20K \texttt{val}.}
  \label{ablation_network}
  \end{subtable}\vspace{2mm}
  \begin{subtable}{0.34\linewidth}
  \centering
  \tablestyle{2.0pt}{1.2}
  \scriptsize
  \begin{tabular}{ccc|ccc}
  Prediction-KD & Pod-KD & Query-KD &  1-100 & 101-150 & \textit{all} \\
  \shline
  &  &  & 0.7 & 2.7 & 1.2\\
  \checkmark &  &  & 35.5 & 15.1 & 28.7\\
  \checkmark &  & \checkmark & 41.3 & 17.9 & 33.5\\
  \checkmark & \checkmark &  & 38.7 & 15.6 & 31.0\\
  \checkmark & \checkmark & \checkmark & 46.2 & 17.6 & 36.7\\
  \end{tabular}
  \caption{\textbf{Query-KD and Prediction-KD.} Experiments are conducted for \textit{100-5} setting on ADE20k \texttt{val}.
  }
  \label{tab:ablation:maskformer:b}
  \end{subtable}\hspace{2mm}
  \begin{subtable}{0.25\linewidth}
  \centering
  \tablestyle{3pt}{1.2}
  \scriptsize
  \begin{tabular}{c|ccc}
  Query Frozen & 1-100 & 101-150 & \textit{all} \\
  \shline
    & 44.2 & 17.3 & 35.2\\
  \checkmark & 46.2  & 17.6  & 36.7 \\
  
  \multicolumn{4}{c}{~}\\
  \end{tabular}
  \caption{\textbf{Frozen for Old Query Embedding.} Experiments are conducted on \textit{100-5} (6 steps) overlapped setting for semantic segmentation~(mIoU) task on ADE20k \texttt{val}.
  }
  \label{ablation_DQ}
  \end{subtable}
  \hspace{2mm}
  \begin{subtable}{0.32\linewidth}
  \centering
  \tablestyle{3pt}{1.2}
  \scriptsize
  \begin{tabular}{c|c|cc|c}
  100-5 & Queries & Params & FPS  &\textit{mIoU} \\
  \shline
  Step 0~(\textit{1-100})& 100 & 43.915 & 11.0 & 32.2\\
  Step 1~(\textit{1-110})& 110  & 43.916 & 10.8 & 32.3\\
  Step 5~(\textit{1-150})& 150 & 43.920 & 9.7 & 36.7\\
  
  \multicolumn{5}{c}{~}\\
  \end{tabular}
  \caption{\textbf{Increasing Params for Dynamic Query.} Experiments are conducted on \textit{100-5} setting for semantic segmentation~(mIoU) task on ADE20K \texttt{val}.
  }
  \label{ablation_param}
  \end{subtable}
  \hspace{2mm}

  \caption{\textbf{CiSDQ ablations.} We perform ablations on ADE20k \texttt{val} for continual semantic segmentation task.
  }
  
  \label{ablation_CiSDQ}
\end{table*}

\begin{table*}[t]
    \centering
    \small 
    \setlength{\tabcolsep}{1mm}

\centering
\setlength{\tabcolsep}{1.3mm}
\footnotesize
\begin{tabular}{ l||c||c||ccc||cccccc } 
\toprule
\multirow{2}*{\textbf{Class}}&
\multirow{2}*{\textbf{Train Set}}&
\multirow{2}*{\textbf{Test Set}} &  
\multicolumn{3}{c||}{\textbf{ADE 50-50 (3 steps,mIoU/\%)}} &
\multicolumn{6}{c}{\textbf{ADE 100-10 (6 steps,mIoU/\%)}} \\
 &  & &  $0$-th step& $1$-th step& $2$-th step& $0$-th step& $1$-th step& $2$-th step& $3$-th step& $4$-th step& $5$-th step\\ 
\midrule
Person &  ADE20K  & VOC 2012 & 
76.7 & 76.4 & 75.6 &
77.2 & 76.3 & 76.7 & 76.8  & 75.5 & 75.0 \\
Car &  ADE20K  & VOC 2012 
& 74.1 &  75.5 & 75.1
& 75.6 & 75.8  &  75.7 &  78.6  & 79.4  & 79.5 \\
Bus & ADE20K  & VOC 2012 & 0 & 68.7 & 68.2 & 62.0 & 61.3  & 60.1  &  52.3  &  50.0 &  50.8 \\
Boat & ADE20K  & VOC 2012 & 0 & 33.4 & 37.7 &  26.0 & 26.5  & 27.2  & 28.7  & 27.7 &  27.4 \\
\bottomrule
\end{tabular}

    \caption{\textbf{The ablation study for model generalization during continual learning.} We conducted cross-dataset validation to investigate the generalization capability of the model. In the ADE 50-50 (3 steps) setting, the 'bus' and 'boat' categories are considered as new classes, requiring the model to learn them in the $1$-th step.}
    \label{generalization}
\end{table*}

\subsection{Ablation Study}
\textbf{Incremental query queue and independent bipartite match}, as two indispensable points, enable our dynamic query and adaptive background.
Table.~\ref{ablation_DQ1} presents the effect of the two components.
Without the independent match, the incremental query just increases the number of queries, which can not decouple the learning of old and new classes.
Therefore, with the only incremental query, \cisdq achieves a little improvement, around $1.0\%$ mIoU.
By contrast, the combination of two components brings obvious improvement, up to $5.2\%$ mIoU over the baseline.

\textbf{Query-KD and Prediction-KD}, two key components for our Query Guided Knowledge Distillation.
Table.~\ref{tab:ablation:maskformer:b} presents the effect for the two components.
Baseline, without the three KDs, training with new classes directly, presents an unacceptable result $1.2\%$ mIoU.
Prediction-KD, Pod-KD, and Query-KD bring different gains for the final results, \ie{} $27.5\%$, $2.3\%$, $5.7\%$, respectively.
Although Pod-KD damages the representation learning of new classes due to the unbalanced matching in Fig.~\ref{fig:3b}, the positive impact for remaining the knowledge of old classes is not neglectable.
Unlike our Query-Guided KD, Pod-KD forces on the features from the backbone and pixel decoder, thus we adopt it in Equ.~\ref{eq:complete_loss}.
Actually, any KDs will damage the performance of newly added classes.
Compared with Pod-KD, Query-KD shows better performance on newly added classes~(\textit{101-150} classes), where $17.9\%$ $v.s$ $15.6\%$ mIou on 3-\textit{th} and 4-\textit{th} lines.
Meanwhile, Query-KD also gives a competitive performance on old classes, as shown in 
Fig.~\ref{VIS}.

\textbf{Frozen for Query of Old Class.}
Except for knowledge distillation, query embedding frozen also help to remain the performance of old classes, as shown in Table.~\ref{ablation_DQ}. 
%
%
With query frozen, the performance of old classes (\textit{1-100} classes) shows an obvious improvement with $1.5\%$ mIoU.
\begin{figure*}[t]
    \includegraphics[width=0.99\linewidth]{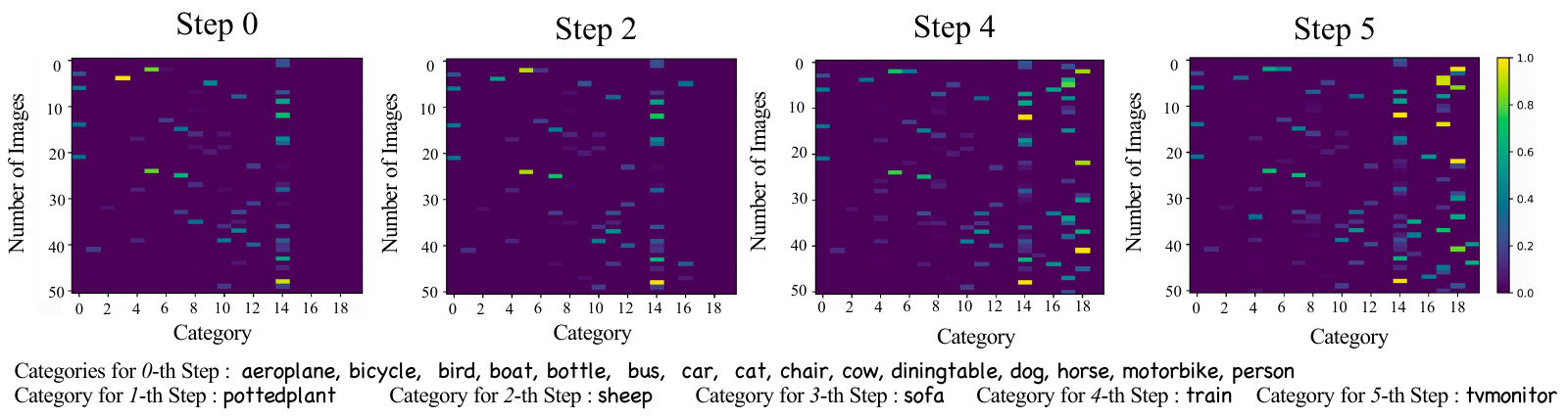}
	\caption{\textbf{Visualization of the output tensor of the decoder for 50 randomly sampled images on the VOC 2012 dataset.}
    Throughout the continual learning process, the model exhibits activations for novel classes, and there is no forgetting of the feature for old classes.
    }
\label{VIS_1}
\end{figure*}

\textbf{Computation Cost from Increasing Query Embedding.}
We also study the extra computation cost from increasing queries in Table.~\ref{ablation_param}.
For \textit{100-5} (11 steps) setting, the number of queries will increase from 100 to 150 to represent the newly added classes.
And the increased 50 queries only bring around $0.005$ M extra parameters, while the performance shows an obvious improvement.
Compared with Deeplab V3 based methods, mask2former-based framework~(ours) shows a big advantage with low computation cost~(fewer parameters, smaller flops, and faster speed) and better performance.

\textbf{Base Framework.} Table.~\ref{ablation_network} gives a comparison for the two different base networks.
Mask2former is a better choice for us to apply continual learning to image segmentation, while the parameters, flops, and FPS are all acceptable.

\textbf{Comparison between PLOP.} Fig.~\ref{VIS123} presents the comparison between PLOP~\cite{(PLOP)douillard2020plop} and \cisdq for semantic segmentation task on ADE 20k 100-5 setting.
As for the learning of newly added classes, \cisdq shows better performance over PLOP, which is mainly caused by two aspects: 1) Different queries decouple the learning of new classes and remaining old knowledge; 2) Query-Guided KD performs a more precise distillation matching without damaging the data distribution of new classes.
Similarly, \cisdq presents less performance loss for old classes during the continual learning of newly added classes process.

\textbf{Ablation Study for Model Generalization.} 
Table~\ref{generalization} present the ablation study for model generalization during continual learning.
We conducted cross-dataset evaluation to study the model's generalization.
Due to variations in the number and distribution of categories across different datasets, such as the inclusion of the 'refrigerator' category in ADE 20K, which is not present in the VOC 2012 dataset, we opted to explore specific categories.
In fact, the generalization variations of the model differ for different categories during the continual learning process.
For example, the generalization of the 'person' category tends to decrease, while the 'car' category shows improvement.
We attribute these changes to a balance between the forgetting of knowledge related to old classes and the decrease in false positives.
In the continual learning process, the forgetting of knowledge related to old classes is a recognized challenge that can lead to a decline in the performance of these classes.
However, as the number of categories increases during continual learning, it can also assist an old class in excluding certain regions that are prone to false positives. 
As a result, different categories and experimental settings may yield varied results. 
Overall, the generalization remains stable. It is essential to emphasize that continual learning is not aimed at enhancing the model's generalization but rather enabling the model to dynamically update its knowledge and extend its ability to segment new classes.

\textbf{Visualization of the output tensor.}
In order to gain a deeper understanding of the continual learning process and the changes in the model, we randomly selected 50 images from the VOC 2012 test set.
Then we extracted and visualized the multiplication of the pixel decoder and transformer decoder outputs.
Given 50 images, we obtain corresponding tensor outputs of size $50 \times 20 \times w \times h$, where $20$ refers to the number of category (Excluding the background category).
For ease of visualization and examining activations for each category, we first use average pooling to transform the tensor from $50 \times 20 \times w \times h$ to $50 \times 20 \times 1 \times 1$. 
Subsequently, we normalize the tensor and visualize the results as shown in Figure~\ref{VIS_1}.
At $0$-th step, it is evident that the tensor lacks activations for the last five classes, such as 'pottedplant' and 'sheep.'
However, as continual learning progresses and the model learns new classes, the tensor exhibits activations for the new classes, allowing for predictions of these novel categories.

\section{Limitations and Future Works}
Although CisDQ can achieve good performance on continual semantic segmentation (CSS), its performance on continual instance segmentation (CIS) is still not satisfactory.
Compared to CSS, CIS requires more queries as instances from the same class can only be distinguished by different queries.
In our future work, we will explore a way to assign queries to the old and novel classes more efficiently, so that classes from each incremental step can get enough queries to achieve high performance on CIS. 

\section{Conclusion}
In this paper, we propose a simple, yet 
effective Continual Image Segmentation method with incremental Dynamic Query, named \cisdq, which 
decouples the representation learning of both old and new knowledge with lightweight query embedding.
\cisdq  also includes a novel class/instance-aware Query Guided Knowledge Distillation strategy to overcome catastrophic forgetting by capturing the inter-class diversity and intra-class identity.
Meanwhile, we introduce the continual learning to instance segmentation task, which is more challenging.
Experimental results show that our \cisdq achieves SOTA performance, with up to $10\%$ mIoU improvement over previous methods.


\section{Acknowledgements} This work is supported by the National Key Research and Development Program of China (\# 2022YFC3602601), and the Key Research and Development Program of Zhejiang Province of China (\# 2021C02037).
M. Shou's participation was supported by Mike Zheng Shou's Start-Up Grant from NUS and the Ministry of Education, Singapore, under the Academic Research Fund Tier 1 (FY2022).

\bibliographystyle{IEEEtran}
\bibliography{IEEEabrv,egbib}  

\begin{thebibliography}{10}
\providecommand{\url}[1]{#1}
\csname url@samestyle\endcsname
\providecommand{\newblock}{\relax}
\providecommand{\bibinfo}[2]{#2}
\providecommand{\BIBentrySTDinterwordspacing}{\spaceskip=0pt\relax}
\providecommand{\BIBentryALTinterwordstretchfactor}{4}
\providecommand{\BIBentryALTinterwordspacing}{\spaceskip=\fontdimen2\font plus
\BIBentryALTinterwordstretchfactor\fontdimen3\font minus \fontdimen4\font\relax}
\providecommand{\BIBforeignlanguage}[2]{{%
\expandafter\ifx\csname l@#1\endcsname\relax
\typeout{** WARNING: IEEEtran.bst: No hyphenation pattern has been}%
\typeout{** loaded for the language `#1'. Using the pattern for}%
\typeout{** the default language instead.}%
\else
\language=\csname l@#1\endcsname
\fi
#2}}
\providecommand{\BIBdecl}{\relax}
\BIBdecl

\bibitem{(deeplabv3)chen2017rethinking}
L.-C. Chen, G.~Papandreou, F.~Schroff, and H.~Adam, ``Rethinking atrous convolution for semantic image segmentation,'' \emph{arXiv preprint arXiv:1706.05587}, 2017.

\bibitem{cheng2022masked}
B.~Cheng, I.~Misra, A.~G. Schwing, A.~Kirillov, and R.~Girdhar, ``Masked-attention mask transformer for universal image segmentation,'' in \emph{Proceedings of the IEEE/CVF Conference on Computer Vision and Pattern Recognition}, 2022, pp. 1290--1299.

\bibitem{(PLOP)douillard2020plop}
A.~Douillard, Y.~Chen, A.~Dapogny, and M.~Cord, ``Plop: Learning without forgetting for continual semantic segmentation,'' \emph{arXiv preprint arXiv:2011.11390}, 2020.

\bibitem{zhang2022representation}
C.-B. Zhang, J.-W. Xiao, X.~Liu, Y.-C. Chen, and M.-M. Cheng, ``Representation compensation networks for continual semantic segmentation,'' in \emph{Proceedings of the IEEE/CVF Conference on Computer Vision and Pattern Recognition}, 2022, pp. 7053--7064.

\bibitem{cha2021ssul}
S.~Cha, Y.~Yoo, T.~Moon \emph{et~al.}, ``Ssul: Semantic segmentation with unknown label for exemplar-based class-incremental learning,'' \emph{Advances in Neural Information Processing Systems}, vol.~34, pp. 10\,919--10\,930, 2021.

\bibitem{robins1995catastrophic}
A.~Robins, ``Catastrophic forgetting, rehearsal and pseudorehearsal,'' \emph{Connection Science}, vol.~7, no.~2, pp. 123--146, 1995.

\bibitem{french1999catastrophic}
R.~M. French, ``Catastrophic forgetting in connectionist networks,'' \emph{Trends in cognitive sciences}, vol.~3, no.~4, pp. 128--135, 1999.

\bibitem{wu2022end}
W.~Wu, D.~Zhang, Y.~Fu, C.~Shen, H.~Zhou, Y.~Cai, and P.~Luo, ``End-to-end video text spotting with transformer,'' \emph{arXiv preprint arXiv:2203.10539}, 2022.

\bibitem{wu2021bilingual}
W.~Wu, D.~Zhang, Y.~Cai, S.~Wang, J.~Li, Z.~Li, Y.~Tang, and H.~Zhou, ``A bilingual, openworld video text dataset and end-to-end video text spotter with transformer,'' in \emph{Thirty-fifth Conference on Neural Information Processing Systems Datasets and Benchmarks Track (Round 2)}, 2021.

\bibitem{zhao2023explore}
Y.~Zhao, Y.~Cai, W.~Wu, and W.~Wang, ``Explore faster localization learning for scene text detection,'' in \emph{2023 IEEE International Conference on Multimedia and Expo (ICME)}.\hskip 1em plus 0.5em minus 0.4em\relax IEEE, 2023, pp. 156--161.

\bibitem{gong2022curiosity}
C.~Gong, Z.~Yang, Y.~Bai, J.~Shi, A.~Sinha, B.~Xu, D.~Lo, X.~Hou, and G.~Fan, ``Curiosity-driven and victim-aware adversarial policies,'' in \emph{Proceedings of the 38th Annual Computer Security Applications Conference}, 2022, pp. 186--200.

\bibitem{allili2010image}
M.~S. Allili, D.~Ziou, N.~Bouguila, and S.~Boutemedjet, ``Image and video segmentation by combining unsupervised generalized gaussian mixture modeling and feature selection,'' \emph{IEEE Transactions on Circuits and Systems for Video Technology}, vol.~20, no.~10, pp. 1373--1377, 2010.

\bibitem{salgado2000efficient}
L.~Salgado, N.~Garcia, J.~M. Menendez, and E.~Rendon, ``Efficient image segmentation for region-based motion estimation and compensation,'' \emph{IEEE transactions on circuits and systems for video technology}, vol.~10, no.~7, pp. 1029--1039, 2000.

\bibitem{ida1995image}
T.~Ida and Y.~Sambonsugi, ``Image segmentation using fractal coding,'' \emph{IEEE transactions on circuits and systems for video technology}, vol.~5, no.~6, pp. 567--570, 1995.

\bibitem{lu2007binary}
H.~Lu, J.~C. Woods, and M.~Ghanbari, ``Binary partition tree for semantic object extraction and image segmentation,'' \emph{IEEE Transactions on Circuits and Systems for Video Technology}, vol.~17, no.~3, pp. 378--383, 2007.

\bibitem{sun2003semiautomatic}
S.~Sun, D.~R. Haynor, and Y.~Kim, ``Semiautomatic video object segmentation using vsnakes,'' \emph{IEEE Transactions on Circuits and Systems for Video Technology}, vol.~13, no.~1, pp. 75--82, 2003.

\bibitem{long2015fully}
J.~Long, E.~Shelhamer, and T.~Darrell, ``Fully convolutional networks for semantic segmentation,'' in \emph{Proceedings of the IEEE conference on computer vision and pattern recognition}, 2015, pp. 3431--3440.

\bibitem{arnab2016higher}
A.~Arnab, S.~Jayasumana, S.~Zheng, and P.~H. Torr, ``Higher order conditional random fields in deep neural networks,'' in \emph{European conference on computer vision}.\hskip 1em plus 0.5em minus 0.4em\relax Springer, 2016, pp. 524--540.

\bibitem{zheng2015conditional}
S.~Zheng, S.~Jayasumana, B.~Romera-Paredes, V.~Vineet, Z.~Su, D.~Du, C.~Huang, and P.~H. Torr, ``Conditional random fields as recurrent neural networks,'' in \emph{Proceedings of the IEEE international conference on computer vision}, 2015, pp. 1529--1537.

\bibitem{zhang2020causal}
D.~Zhang, H.~Zhang, J.~Tang, X.-S. Hua, and Q.~Sun, ``Causal intervention for weakly-supervised semantic segmentation,'' \emph{Advances in Neural Information Processing Systems}, vol.~33, pp. 655--666, 2020.

\bibitem{zhang2021self}
------, ``Self-regulation for semantic segmentation,'' in \emph{Proceedings of the IEEE/CVF International Conference on Computer Vision}, 2021, pp. 6953--6963.

\bibitem{chen2014semantic}
L.-C. Chen, G.~Papandreou, I.~Kokkinos, K.~Murphy, and A.~L. Yuille, ``Semantic image segmentation with deep convolutional nets and fully connected crfs,'' \emph{arXiv preprint arXiv:1412.7062}, 2014.

\bibitem{zhao2017pyramid}
H.~Zhao, J.~Shi, X.~Qi, X.~Wang, and J.~Jia, ``Pyramid scene parsing network,'' in \emph{Proceedings of the IEEE conference on computer vision and pattern recognition}, 2017, pp. 2881--2890.

\bibitem{liu2015parsenet}
W.~Liu, A.~Rabinovich, and A.~C. Berg, ``Parsenet: Looking wider to see better,'' \emph{arXiv preprint arXiv:1506.04579}, 2015.

\bibitem{eigen2015predicting}
D.~Eigen and R.~Fergus, ``Predicting depth, surface normals and semantic labels with a common multi-scale convolutional architecture,'' in \emph{Proceedings of the IEEE international conference on computer vision}, 2015, pp. 2650--2658.

\bibitem{farabet2012learning}
C.~Farabet, C.~Couprie, L.~Najman, and Y.~LeCun, ``Learning hierarchical features for scene labeling,'' \emph{IEEE transactions on pattern analysis and machine intelligence}, vol.~35, no.~8, pp. 1915--1929, 2012.

\bibitem{xie2021segformer}
E.~Xie, W.~Wang, Z.~Yu, A.~Anandkumar, J.~M. Alvarez, and P.~Luo, ``Segformer: Simple and efficient design for semantic segmentation with transformers,'' \emph{Advances in Neural Information Processing Systems}, vol.~34, pp. 12\,077--12\,090, 2021.

\bibitem{cheng2021per}
B.~Cheng, A.~Schwing, and A.~Kirillov, ``Per-pixel classification is not all you need for semantic segmentation,'' \emph{Advances in Neural Information Processing Systems}, vol.~34, pp. 17\,864--17\,875, 2021.

\bibitem{hariharan2014simultaneous}
B.~Hariharan, P.~Arbel{\'a}ez, R.~Girshick, and J.~Malik, ``Simultaneous detection and segmentation,'' in \emph{European conference on computer vision}.\hskip 1em plus 0.5em minus 0.4em\relax Springer, 2014, pp. 297--312.

\bibitem{he2017mask}
K.~He, G.~Gkioxari, P.~Doll{\'a}r, and R.~Girshick, ``Mask r-cnn,'' in \emph{Proceedings of the IEEE international conference on computer vision}, 2017, pp. 2961--2969.

\bibitem{ren2015faster}
S.~Ren, K.~He, R.~Girshick, and J.~Sun, ``Faster r-cnn: Towards real-time object detection with region proposal networks,'' \emph{Advances in neural information processing systems}, vol.~28, 2015.

\bibitem{carion2020end}
N.~Carion, F.~Massa, G.~Synnaeve, N.~Usunier, A.~Kirillov, and S.~Zagoruyko, ``End-to-end object detection with transformers,'' in \emph{European conference on computer vision}.\hskip 1em plus 0.5em minus 0.4em\relax Springer, 2020, pp. 213--229.

\bibitem{Zhao_2023_ICCV}
Y.~Zhao, Q.~Ye, W.~Wu, C.~Shen, and F.~Wan, ``Generative prompt model for weakly supervised object localization,'' in \emph{Proceedings of the IEEE/CVF International Conference on Computer Vision (ICCV)}, October 2023, pp. 6351--6361.

\bibitem{wu2023diffumask}
W.~Wu, Y.~Zhao, M.~Z. Shou, H.~Zhou, and C.~Shen, ``Diffumask: Synthesizing images with pixel-level annotations for semantic segmentation using diffusion models,'' \emph{Proc. Int. Conf. Computer Vision (ICCV 2023)}, 2023.

\bibitem{wu2023datasetdm}
W.~Wu, Y.~Zhao, H.~Chen, Y.~Gu, R.~Zhao, Y.~He, H.~Zhou, M.~Z. Shou, and C.~Shen, ``Datasetdm: Synthesizing data with perception annotations using diffusion models,'' \emph{arXiv preprint arXiv:2308.06160}, 2023.

\bibitem{yu2023contrastive}
D.~Yu, M.~Zhang, M.~Li, F.~Zha, J.~Zhang, L.~Sun, and K.~Huang, ``Contrastive correlation preserving replay for online continual learning,'' \emph{IEEE Transactions on Circuits and Systems for Video Technology}, 2023.

\bibitem{fu2023continual}
X.~Fu, J.~Xiao, Y.~Zhu, A.~Liu, F.~Wu, and Z.-J. Zha, ``Continual image deraining with hypergraph convolutional networks,'' \emph{IEEE Transactions on Pattern Analysis and Machine Intelligence}, 2023.

\bibitem{le2022uifgan}
Z.~Le, J.~Huang, H.~Xu, F.~Fan, Y.~Ma, X.~Mei, and J.~Ma, ``Uifgan: An unsupervised continual-learning generative adversarial network for unified image fusion,'' \emph{Information Fusion}, vol.~88, pp. 305--318, 2022.

\bibitem{zhou2021image}
M.~Zhou, J.~Xiao, Y.~Chang, X.~Fu, A.~Liu, J.~Pan, and Z.-J. Zha, ``Image de-raining via continual learning,'' in \emph{Proceedings of the IEEE/CVF Conference on Computer Vision and Pattern Recognition}, 2021, pp. 4907--4916.

\bibitem{xu2020u2fusion}
H.~Xu, J.~Ma, J.~Jiang, X.~Guo, and H.~Ling, ``U2fusion: A unified unsupervised image fusion network,'' \emph{IEEE Transactions on Pattern Analysis and Machine Intelligence}, vol.~44, no.~1, pp. 502--518, 2020.

\bibitem{yan2020interactive}
R.~Yan, L.~Xie, X.~Shu, and J.~Tang, ``Interactive fusion of multi-level features for compositional activity recognition,'' \emph{arXiv preprint arXiv:2012.05689}, 2020.

\bibitem{yan2020social}
R.~Yan, L.~Xie, J.~Tang, X.~Shu, and Q.~Tian, ``Social adaptive module for weakly-supervised group activity recognition,'' in \emph{Computer Vision--ECCV 2020: 16th European Conference, Glasgow, UK, August 23--28, 2020, Proceedings, Part VIII 16}.\hskip 1em plus 0.5em minus 0.4em\relax Springer, 2020, pp. 208--224.

\bibitem{mccloskey1989catastrophic}
M.~McCloskey and N.~J. Cohen, ``Catastrophic interference in connectionist networks: The sequential learning problem,'' in \emph{Psychology of learning and motivation}.\hskip 1em plus 0.5em minus 0.4em\relax Elsevier, 1989, vol.~24, pp. 109--165.

\bibitem{bang2021rainbow}
J.~Bang, H.~Kim, Y.~Yoo, J.-W. Ha, and J.~Choi, ``Rainbow memory: Continual learning with a memory of diverse samples,'' in \emph{Proceedings of the IEEE/CVF Conference on Computer Vision and Pattern Recognition}, 2021, pp. 8218--8227.

\bibitem{belouadah2019il2m}
E.~Belouadah and A.~Popescu, ``Il2m: Class incremental learning with dual memory,'' in \emph{Proceedings of the IEEE/CVF International Conference on Computer Vision}, 2019, pp. 583--592.

\bibitem{chaudhry2021using}
A.~Chaudhry, A.~Gordo, P.~Dokania, P.~Torr, and D.~Lopez-Paz, ``Using hindsight to anchor past knowledge in continual learning,'' in \emph{Proceedings of the AAAI Conference on Artificial Intelligence}, vol.~35, no.~8, 2021, pp. 6993--7001.

\bibitem{chaudhry2018riemannian}
A.~Chaudhry, P.~K. Dokania, T.~Ajanthan, and P.~H. Torr, ``Riemannian walk for incremental learning: Understanding forgetting and intransigence,'' in \emph{Proceedings of the European Conference on Computer Vision (ECCV)}, 2018, pp. 532--547.

\bibitem{cheraghian2021semantic}
A.~Cheraghian, S.~Rahman, P.~Fang, S.~K. Roy, L.~Petersson, and M.~Harandi, ``Semantic-aware knowledge distillation for few-shot class-incremental learning,'' in \emph{Proceedings of the IEEE/CVF Conference on Computer Vision and Pattern Recognition}, 2021, pp. 2534--2543.

\bibitem{douillard2020podnet}
A.~Douillard, M.~Cord, C.~Ollion, T.~Robert, and E.~Valle, ``Podnet: Pooled outputs distillation for small-tasks incremental learning,'' in \emph{European Conference on Computer Vision}.\hskip 1em plus 0.5em minus 0.4em\relax Springer, 2020, pp. 86--102.

\bibitem{xiang2019incremental}
Y.~Xiang, Y.~Fu, P.~Ji, and H.~Huang, ``Incremental learning using conditional adversarial networks,'' in \emph{Proceedings of the IEEE/CVF International Conference on Computer Vision}, 2019, pp. 6619--6628.

\bibitem{ebrahimi2020adversarial}
S.~Ebrahimi, F.~Meier, R.~Calandra, T.~Darrell, and M.~Rohrbach, ``Adversarial continual learning,'' in \emph{European Conference on Computer Vision}.\hskip 1em plus 0.5em minus 0.4em\relax Springer, 2020, pp. 386--402.

\bibitem{dosovitskiy2020image}
A.~Dosovitskiy, L.~Beyer, A.~Kolesnikov, D.~Weissenborn, X.~Zhai, T.~Unterthiner, M.~Dehghani, M.~Minderer, G.~Heigold, S.~Gelly \emph{et~al.}, ``An image is worth 16x16 words: Transformers for image recognition at scale,'' \emph{arXiv preprint arXiv:2010.11929}, 2020.

\bibitem{xue2022meta}
M.~Xue, H.~Zhang, J.~Song, and M.~Song, ``Meta-attention for vit-backed continual learning,'' in \emph{Proceedings of the IEEE/CVF Conference on Computer Vision and Pattern Recognition}, 2022, pp. 150--159.

\bibitem{ashok2022class}
A.~Ashok, K.~Joseph, and V.~Balasubramanian, ``Class-incremental learning with cross-space clustering and controlled transfer,'' \emph{arXiv preprint arXiv:2208.03767}, 2022.

\bibitem{douillard2022dytox}
A.~Douillard, A.~Ram{\'e}, G.~Couairon, and M.~Cord, ``Dytox: Transformers for continual learning with dynamic token expansion,'' in \emph{Proceedings of the IEEE/CVF Conference on Computer Vision and Pattern Recognition}, 2022, pp. 9285--9295.

\bibitem{boschini2022transfer}
M.~Boschini, L.~Bonicelli, A.~Porrello, G.~Bellitto, M.~Pennisi, S.~Palazzo, C.~Spampinato, and S.~Calderara, ``Transfer without forgetting,'' \emph{arXiv preprint arXiv:2206.00388}, 2022.

\bibitem{huang2021half}
Z.~Huang, W.~Hao, X.~Wang, M.~Tao, J.~Huang, W.~Liu, and X.-S. Hua, ``Half-real half-fake distillation for class-incremental semantic segmentation,'' \emph{arXiv preprint arXiv:2104.00875}, 2021.

\bibitem{yan2021framework}
S.~Yan, J.~Zhou, J.~Xie, S.~Zhang, and X.~He, ``An em framework for online incremental learning of semantic segmentation,'' in \emph{Proceedings of the 29th ACM International Conference on Multimedia}, 2021, pp. 3052--3060.

\bibitem{cermelli2020modeling}
F.~Cermelli, M.~Mancini, S.~R. Bulo, E.~Ricci, and B.~Caputo, ``Modeling the background for incremental learning in semantic segmentation,'' in \emph{Proceedings of the IEEE/CVF Conference on Computer Vision and Pattern Recognition}, 2020, pp. 9233--9242.

\bibitem{zhao2022rbc}
H.~Zhao, F.~Yang, X.~Fu, and X.~Li, ``Rbc: Rectifying the biased context in continual semantic segmentation,'' in \emph{European Conference on Computer Vision}.\hskip 1em plus 0.5em minus 0.4em\relax Springer, 2022, pp. 55--72.

\bibitem{zheng2021continual}
E.~Zheng, Q.~Yu, R.~Li, P.~Shi, and A.~Haake, ``A continual learning framework for uncertainty-aware interactive image segmentation,'' in \emph{Proceedings of the AAAI Conference on Artificial Intelligence}, vol.~35, no.~7, 2021, pp. 6030--6038.

\bibitem{(MiB)cermelli2020modeling}
F.~Cermelli, M.~Mancini, S.~R. Bulo, E.~Ricci, and B.~Caputo, ``Modeling the background for incremental learning in semantic segmentation,'' in \emph{Proceedings of the IEEE/CVF Conference on Computer Vision and Pattern Recognition}, 2020, pp. 9233--9242.

\bibitem{zhang2022mining}
Z.~Zhang, G.~Gao, Z.~Fang, J.~Jiao, and Y.~Wei, ``Mining unseen classes via regional objectness: A simple baseline for incremental segmentation,'' \emph{Advances in Neural Information Processing Systems}, vol.~35, pp. 24\,340--24\,353, 2022.

\bibitem{qiu2022sats}
Y.~Qiu, Y.~Shen, Z.~Sun, Y.~Zheng, X.~Chang, W.~Zheng, and R.~Wang, ``Sats: Self-attention transfer for continual semantic segmentation,'' \emph{arXiv preprint arXiv:2203.07667}, 2022.

\bibitem{ganea2021incremental}
D.~A. Ganea, B.~Boom, and R.~Poppe, ``Incremental few-shot instance segmentation,'' in \emph{Proceedings of the IEEE/CVF Conference on Computer Vision and Pattern Recognition}, 2021, pp. 1185--1194.

\bibitem{nguyen2022ifs}
K.~Nguyen and S.~Todorovic, ``ifs-rcnn: An incremental few-shot instance segmenter,'' in \emph{Proceedings of the IEEE/CVF Conference on Computer Vision and Pattern Recognition}, 2022, pp. 7010--7019.

\bibitem{milletari2016v}
F.~Milletari, N.~Navab, and S.-A. Ahmadi, ``V-net: Fully convolutional neural networks for volumetric medical image segmentation,'' in \emph{2016 fourth international conference on 3D vision (3DV)}.\hskip 1em plus 0.5em minus 0.4em\relax IEEE, 2016, pp. 565--571.

\bibitem{(ILT)michieli2019incremental}
U.~Michieli and P.~Zanuttigh, ``Incremental learning techniques for semantic segmentation,'' in \emph{Proceedings of the IEEE/CVF International Conference on Computer Vision Workshops}, 2019, pp. 0--0.

\bibitem{xiao2023endpoints}
J.-W. Xiao, C.-B. Zhang, J.~Feng, X.~Liu, J.~van~de Weijer, and M.-M. Cheng, ``Endpoints weight fusion for class incremental semantic segmentation,'' in \emph{Proceedings of the IEEE/CVF Conference on Computer Vision and Pattern Recognition}, 2023, pp. 7204--7213.

\bibitem{cermelli2023comformer}
F.~Cermelli, M.~Cord, and A.~Douillard, ``Comformer: Continual learning in semantic and panoptic segmentation,'' in \emph{Proceedings of the IEEE/CVF Conference on Computer Vision and Pattern Recognition}, 2023, pp. 3010--3020.

\bibitem{hou2017deeply}
Q.~Hou, M.-M. Cheng, X.~Hu, A.~Borji, Z.~Tu, and P.~H. Torr, ``Deeply supervised salient object detection with short connections,'' in \emph{Proceedings of the IEEE conference on computer vision and pattern recognition}, 2017, pp. 3203--3212.

\bibitem{liu2010learning}
T.~Liu, Z.~Yuan, J.~Sun, J.~Wang, N.~Zheng, X.~Tang, and H.-Y. Shum, ``Learning to detect a salient object,'' \emph{IEEE Transactions on Pattern analysis and machine intelligence}, vol.~33, no.~2, pp. 353--367, 2010.

\bibitem{(LwF)LiHoiem16}
Z.~Li and D.~Hoiem, ``Learning without forgetting,'' \emph{IEEE Transactions on Pattern Analysis and Machine Intelligence}, vol.~40, no.~12, pp. 2935--2947, 2017.

\bibitem{(voc)everingham2010pascal}
M.~Everingham, L.~Van~Gool, C.~K. Williams, J.~Winn, and A.~Zisserman, ``The pascal visual object classes (voc) challenge,'' \emph{International journal of computer vision}, vol.~88, no.~2, pp. 303--338, 2010.

\bibitem{(ade)zhou2017scene}
B.~Zhou, H.~Zhao, X.~Puig, S.~Fidler, A.~Barriuso, and A.~Torralba, ``Scene parsing through ade20k dataset,'' in \emph{Proceedings of the IEEE conference on computer vision and pattern recognition}, 2017, pp. 633--641.

\bibitem{cermelli2020modelingthebackground}
F.~Cermelli, M.~Mancini, S.~R. Bulo, E.~Ricci, and B.~Caputo, ``Modeling the background for incremental learning in semantic segmentation,'' in \emph{Proceedings of the IEEE/CVF Conference on Computer Vision and Pattern Recognition}, 2020, pp. 9233--9242.

\bibitem{lin2014microsoft}
T.-Y. Lin, M.~Maire, S.~Belongie, J.~Hays, P.~Perona, D.~Ramanan, P.~Doll{\'a}r, and C.~L. Zitnick, ``Microsoft coco: Common objects in context,'' in \emph{European conference on computer vision}.\hskip 1em plus 0.5em minus 0.4em\relax Springer, 2014, pp. 740--755.

\end{thebibliography}

\end{document}